%% file: cvpr2023_main.tex
\documentclass[10pt,twocolumn,letterpaper]{article}

\usepackage{iccv}
\usepackage{times}

\usepackage{graphicx}
\usepackage{amsmath}
\usepackage{amssymb}
\usepackage{booktabs}

\usepackage{url}            
\usepackage{booktabs}       
\usepackage{amsfonts}       
\usepackage{nicefrac}       
\usepackage{microtype}      
\usepackage{xcolor}         
\usepackage[ruled,vlined,linesnumbered]{algorithm2e}

\usepackage[accsupp]{axessibility}

\usepackage{lipsum}

\usepackage{makecell}
\usepackage{bbm}
\usepackage{wrapfig}
\usepackage{multirow}
\usepackage{subcaption}
\usepackage[toc,page,header]{appendix}
\usepackage{minitoc}
\usepackage{algpseudocode}
\usepackage{dsfont}
\usepackage{pifont}
\usepackage{comment}

\usepackage{colortbl}

\usepackage[pagebackref, breaklinks, colorlinks,
            citecolor=citecolor, linkcolor=linkcolor, bookmarks=false]{hyperref}
\definecolor{citecolor}{HTML}{0071BC}
\definecolor{linkcolor}{HTML}{ED1C24}

\usepackage[capitalize]{cleveref}
\crefname{section}{Sec.}{Secs.}
\Crefname{section}{Section}{Sections}
\Crefname{table}{Table}{Tables}
\crefname{table}{Tab.}{Tabs.}

\def\iccvPaperID{6078}

\iccvfinalcopy 
\ificcvfinal\fi

\providecommand\citep{}
\renewcommand\citep{\cite}

\providecommand\citet{}
\renewcommand\citet{\cite}

\newcommand{\cmark}{\ding{51}}%
\newcommand{\xmark}{\ding{55}}%

\def\etal{\emph{et al.}}
\def\eg{\emph{e.g.}}
\def\ie{\emph{i.e.}}
\def\wrt{w.r.t.}

\newcommand{\revision}[1]{{\color{black}{#1}}}
\newcommand{\iclrrevision}[1]{{\color{black}{#1}}}
\newcommand{\iclrrebuttal}[1]{{\color{black}{#1}}}

\newcommand{\cvprrevision}[1]{{\color{black}{#1}}}
\newcommand{\iccvrevision}[1]{{\color{black}{#1}}}
\definecolor{baselinecolor}{gray}{.9}

\newcommand{\uline}[1]{\underline{#1}}

\definecolor{shaded_gray}{gray}{.9}

\begin{document}

\title{Learning Semi-supervised Gaussian Mixture Models \\for Generalized Category Discovery}

\author{Bingchen Zhao\textsuperscript{\textdagger} \qquad 
Xin Wen\textsuperscript{\textdaggerdbl} \qquad
Kai Han\textsuperscript{\textdaggerdbl}\footnotemark[1]\\
\textsuperscript{\textdagger}University of Edinburgh \quad 
\textsuperscript{\textdaggerdbl}The University of Hong Kong\\
{\tt\small bingchen.zhao@ed.ac.uk \quad wenxin@eee.hku.hk \quad kaihanx@hku.hk}
}
\maketitle
\thispagestyle{empty}
\renewcommand{\thefootnote}{\fnsymbol{footnote}}
\footnotetext[1]{Corresponding author.}

\begin{abstract}
In this paper, we address the problem of generalized category discovery (GCD), \ie, given a set of images where part of them are labelled and the rest are not, the task is to automatically cluster the images in the unlabelled data, leveraging the information from the labelled data, while the unlabelled data contain images from the labelled classes and also new ones. GCD is similar to semi-supervised learning (SSL) but is more realistic and challenging, as SSL assumes all the unlabelled images are from the same classes as the labelled ones. 
We also do not assume the class number in the unlabelled data is known a-priori, making the GCD problem even harder. 
To tackle the problem of GCD without knowing the class number, we propose an EM-like framework that alternates between representation learning and class number estimation. We propose a semi-supervised variant of the Gaussian Mixture Model (GMM) with a stochastic splitting and merging mechanism to dynamically determine the prototypes by examining the cluster compactness and separability. With these prototypes, we leverage prototypical contrastive learning for representation learning on the partially labelled data subject to the constraints imposed by the labelled data. Our framework alternates between these two steps until convergence. The cluster assignment for an unlabelled instance can then be retrieved by identifying its nearest prototype. We comprehensively evaluate our framework on both generic image classification datasets and challenging fine-grained object recognition datasets, achieving state-of-the-art performance. 
\end{abstract}


\input{cvpr/intro_cvpr.tex}

\input{cvpr/related_cvpr.tex}

\input{cvpr/method_cvpr.tex}

\input{cvpr/exp_cvpr.tex}
\input{cvpr/conclusion_cvpr.tex}

{\small
\bibliographystyle{ieee_fullname}
\bibliography{iclr2023_conference}
}

\clearpage

\appendix
\onecolumn

\input{cvpr/supp_cvpr.tex}

\end{document}

%% file: cvpr/intro_cvpr.tex
\section{Introduction}
\label{sec:intro}

\begin{figure}[t]
    \centering
    \includegraphics[width=\linewidth]{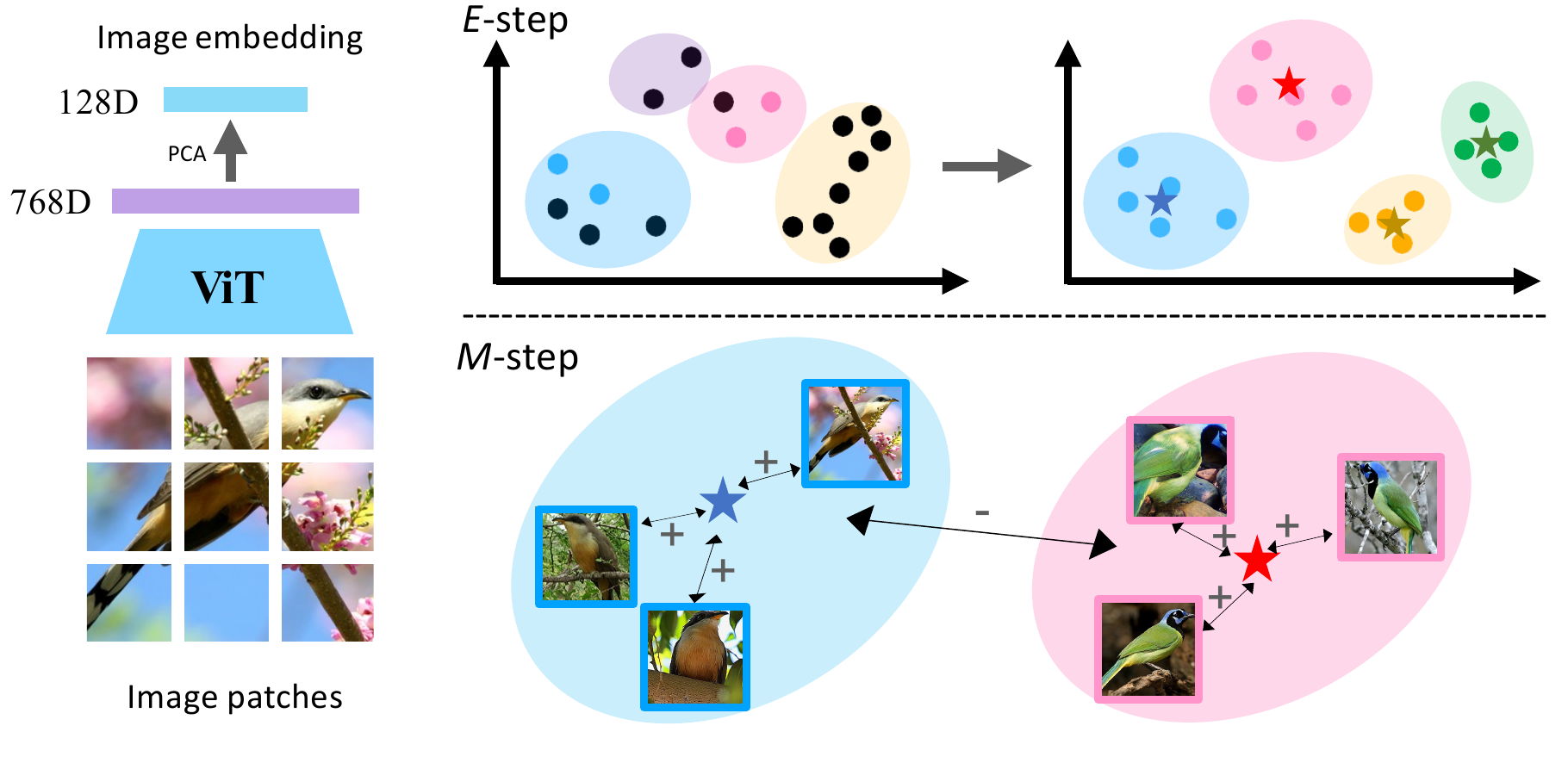}
    \caption{\textbf{Overview of our proposed EM-like framework.} The input images are fed into a ViT-B model to obtain a 768-dimensional feature vector, then the feature vector will be projected to a lower dimensional space using the projection calculated from PCA. We perform class number estimation and representation learning in this projected space. In the E-step, we use a semi-supervised GMM that can split separable clusters and merge cluttered clusters to estimate the class number and prototypes, which will be used in the M-step of representation learning with prototypical contrastive learning.}
    \label{fig:framework}
    \vspace{-6pt}
\end{figure}

The success of deep learning is driven by the availability of large-scale data with human annotations. 
Given enough annotated data, deep learning models are able to surpass human-level performance on many important computer vision tasks such as image classification~\citep{resnet}.
But the cost of collecting a large annotated dataset is not always affordable, and it is also not possible to annotate all new classes emerging from the real world.
Thus, designing models that can learn to deal with large-scale unlabelled data in the open world is of great value and importance.
Semi-supervised learning (SSL)~\citep{oliver2018realistic} is proposed as a solution to learn a model on both labelled data and unlabelled data, with many works achieving promising performance~\citep{berthelot2019mixmatch,tarvainen2017mean,sohn2020fixmatch}.
However, SSL assumes that labelled instances are provided for all object classes in the unlabelled data.
The novel category discovery (NCD) task is introduced~\citep{Han2019learning,han21autonovel} to automatically discover novel classes by transferring the knowledge learned from the labelled instances of known classes, assuming the unlabelled data only contain instances from new classes.
Generalized category discovery (GCD)~\citep{vaze22generalized} further relaxes the assumption in NCD, and tackles a more generalized setting where the unlabelled data contains instances from both known and novel categories.
Existing methods for NCD~\citep{Han2019learning,han21autonovel,zhao21novel,zhong2021neighborhood,zhong2021openmix,fini2021unified,jia2021joint} and GCD~\citep{vaze22generalized,fei2022xcon,wen2022simple,sun2023opencon,zhang2022promptcal,zhao2023incremental,liu2023large} learn the representation and cluster assignment assuming the class number is known a priori~\citep{zhao21novel,jia2021joint,zhong2021neighborhood,fini2021unified,zhong2021openmix} or precomputed~\citep{Han2019learning,vaze22generalized}. In practice, the number of categories in the unlabelled data is often unknown, while precomputing the class number without taking the representation learning into consideration is likely to lead to a sub-optimal solution.


In this paper, we argue that representation learning and the estimation of class numbers should be considered together and could reinforce each other, \ie, a strong representation could help a more accurate estimation of the class numbers, and an accurate class number could help learn a better feature representation.
To this end, we propose a unified EM-like framework that alternates between feature representation learning and class number estimation where the E-step is aimed at automatically estimating a proper class number and a set of class prototypes in the unlabelled data and the M-step is aimed at learning better representation with the class number and class prototypes estimated.  
In particular, we propose using a prototype contrastive representation learning~\citep{li2020prototypical} method for GCD, which requires a set of prototypes to serve as anchors for representation learning. Prototypical contrastive learning~\citep{li2020prototypical} is developed for unsupervised representation learning to generalize to different tasks, where the prototypes are obtained by over-clustering  the dataset with one or multiple given prototype numbers, using non-parametric clustering algorithms like $k$-means.
Instead, to handle the problem of GCD, we propose to estimate the prototype number and prototypes automatically and simultaneously. To do so, we introduce a semi-supervised variant of the Gaussian Mixture Model (GMM) with a stochastic splitting and merging mechanism to determine the most suitable clusters based on current representation. These clusters can then be used to form prototypes to facilitate contrastive representation learning. Our framework alternates between the E- and M-step until converging to achieve robust representation and reliable category estimation. After learning, the cluster assignment for an unlabelled instance, either from known or novel classes, can be retrieved by finding the nearest prototypes. 
Thus we name our framework as \textbf{GPC}: \textbf{G}aussian mixture model for generalized category discovery with \textbf{P}rotypical \textbf{C}ontrastive learning.



Our contributions in this paper are as follows:
(1) We demonstrate that in generalized category discovery, the class number estimation and representation learning can reinforce each other in the learning process. Strong representations can give a better estimation of the class number, and vice versa. 
(2) We propose an EM-like framework that alternates between prototype estimation with a variant of GMM (E-step) and representation learning based on prototypical contrastive learning (M-step).
(3) We introduce a semi-supervised variant of GMM with a stochastic splitting and merging mechanism to allow dynamic change of the prototypes by examining the cluster compactness and separability based on the Metropolis-Hastings ratio~\citep{hastings1970monte}. 
(4) We comprehensively evaluated our framework on both the generic image classification benchmark, including CIFAR10, CIFAR100, ImageNet-100, and the challenging fine-grained Semantic Shifts Benchmark suite, which includes CUB-200, Stanford-Cars, and FGVC-aircrafts, achieving the state-of-the-art results.




%% file: cvpr/related_cvpr.tex
\section{Related work}
\label{sec:related}


\textit{Novel category discovery }(NCD) is first formalized in DTC~\citep{Han2019learning}, where the task is to discover new categories leveraging the knowledge of a set of labelled categories.
Earlier methods like MCL~\citep{hsu2017learning} and KCL~\citep{hsu2019multi} for generalized transfer learning can also be applied to this problem.
RankStat~\citep{rankstat,han21autonovel} shows that this task benefits from self-supervised pretraining and proposes a method to transfer knowledge from the labelled data to the unlabelled data using ranking statistics.
NCL~\citep{zhong2021neighborhood} and \citet{jia2021joint} adopt contrastive learning for novel category discovery.
OpenMix~\citep{zhong2021openmix} shows that mixing up labelled and unlabelled data can help avoid the representation from overfitting to the labelled categories.
\citet{zhao21novel} propose a dual ranking statistics framework to focus on the local visual cues to improve the performance on fine-grained classification benchmarks.
UNO~\citep{fini2021unified} introduces a unified cross-entropy loss that enables the model to be jointly trained on unlabelled and labelled data.
Most recently, generalized category discovery (GCD) is introduced in~\citet{vaze22generalized} to extend NCD to a more open-world setting where unlabelled instances can come from both labelled and unlabelled categories. 
Concurrent work ORCA~\citep{cao21orca} also tackles a similar setting as GCD, termed open-world semi-supervised learning.
Later works~\cite{fei2022xcon,wen2022simple,Pu_2023_CVPR,zhang2022promptcal} extends on the GCD setting and proposed more effective methods, some works also explored the incremental setting of the GCD problem~\cite{zhao2023incremental,incd2022}
Despite the advance in this setting, most methods still assume that the novel class number is known \textit{a priori}, which is often not the case in the real world.
To address this problem, DTC~\citep{Han2019learning} and GCD~\citep{vaze22generalized} precompute the number of novel classes using a semi-supervised $k$-means algorithm, without considering the representation learning.
In this paper, we demonstrate that class number estimation and representation learning can be jointly considered to mutually benefit each other.

\textit{Contrastive learning}~\citep{chen2020simple,chen2020improved,he2019moco,cui2022discriminability,zhu2021improving} (CL) has been shown very effective for representation learning in a self-supervised manner, using the instance discrimination pretext~\citep{wu2018unsupervised} as the learning objective. 
The instance discrimination task learns a representation by pulling positive samples from the augmentations of the same images closer and pushing negative samples from different images apart in the embedding space.
Instead of contrasting over all instances in a mini-bath, prototypical contrastive learning (PCL)~\citep{li2020prototypical} proposes to contrast the features with a set of prototypes which can provide a higher level abstraction of dataset than instances and has been shown to be more data efficient without the need of large batch size. 
Though PCL is developed for unsupervised representation learning, if the prototypes are viewed as cluster centers, it can be leveraged in the partially supervised setting of GCD for representation learning to better fit the GCD task of partitioning data into different clusters. Thus, in this paper, we adopt PCL to fit the GCD setting for representation learning in which the downstream clustering task is directly considered.

\textit{Semi-supervised learning} (SSL) has been a long standing research topic which many effective method proposed~\citep{rebuffi2020semi,sohn2020fixmatch,berthelot2019mixmatch,laine2016temporal,tarvainen2017mean}.
In SSL, the labelled and the unlabelled data are assumed to come from the same set of classes, and the task is to learn a classification model that can take advantage of both labelled and unlabelled data.
Consistency-based methods are among the most effective methods for SSL, such as Mean-teacher~\citep{tarvainen2017mean}, MixMatch~\citep{berthelot2019mixmatch}, and FixMatch~\citep{sohn2020fixmatch}.
Self-supervised representation learning also shows to be helpful for SSL because it can provide a strong representation~\citep{zhai2019s4l,rebuffi2020semi}.
\iccvrevision{
Recent works extend semi-supervised learning by relaxing the assumption of exactly the same classes in the labeled and unlabelled data~\cite{saito2021openmatch,huang2021trash,yu2020multi}, but their focus is improving the performance of the labeled categories without discovering novel categories in the unlabeled set.
}

\textit{Unsupervised clustering} has been studied for decades, and there are many existing classical approaches~\citep{macqueen1967some_kmeans,ester1996density,comaniciu2002mean_shift} as well as deep learning based approaches~\citep{xie2016unsupervised,rebuffi21lsdc,ghasedi2017deep}. 
Recently, DeepDPM~\citep{ronen2022deepdpm} is proposed to automatically determine the number of clusters for a given dataset by adopting a similar split/merge framework that changes the inferred number of clusters.
However, due to the unsupervised nature of these methods, there is no prior or supervision over how a cluster should be formed, thus multiple equally valid clustering results following different clustering criteria can be produced.
Thus, directly applying unsupervised clustering methods to the task of generalized category discovery is not feasible, as we would want the model to use one unique clustering criteria implicitly given by the labelled data.

%% file: cvpr/method_cvpr.tex
\section{Method}
\label{sec:method}



Given a collection of partially labelled data, $\mathcal{D} = \mathcal{D}^l \cup \mathcal{D}^u$, where $\mathcal{D}^l = \{(x_i, y^l_i)\} \in \mathcal{X} \times \mathcal{Y}_{l}$ is labelled, $\mathcal{D}^u = \{x_i, y^u_i\} \in \mathcal{X} \times \mathcal{Y}_{u}$ is unlabelled, and $\mathcal{Y}_{l} \subset \mathcal{Y}_{u}$, Generalized category discovery (GCD) aims at automatically assign labels for the unlabelled instances in $\mathcal{D}^u$, by transferring knowledge acquired from $\mathcal{D}^l$. 
Let the category number in $\mathcal{D}^l$ be $K^l = |\mathcal{Y}_l|$ and that in $\mathcal{D}^u$ be $K^u = |\mathcal{Y}_u|$. The number of new categories $\mathcal{D}^u$ is then $K^n = |\mathcal{Y}_u \setminus \mathcal{Y}_l| = K^u - K^l$. Though $K^l$ can be accessed from the labelled data, we do not assume $K^n$ or $K^u$ to be known. This is a realistic setting to reflect the real open world, where we often have access to some labelled data, but in the unlabelled data, we also have instances from unseen new categories. 

The key challenges for GCD are representation learning, category number estimation, and label assignment.
Existing methods for NCD and GCD~\citep{Han2019learning,vaze22generalized} deal with these three challenges independently.
However, we believe they are inherently linked with each other. 
Label assignment depends on representation and category number estimation. 
A good class number estimation can facilitate representation learning, thus better label assignment, and vice versa.
Thus, in this paper, we aim to jointly handle these challenges in the learning process for a more reliable GCD. 

To this end, we propose a unified EM-like framework that alternates between representation learning and class number estimation, while the label assignment turns out to be a by-product during class number estimation. In the E-step, we introduce a semi-supervised variant of the Gaussian Mixture Model (GMM) to estimate the class numbers by dynamically splitting separable clusters and merging cluttered clusters based on current representation, forming a set of class prototypes for both seen and unseen classes, and in the M-step, we train the model to produce discriminative representation by prototypical contrastive learning using the cluster centers from the GMM prototypes derived from the E-step during class number estimation.
After training, the class assignment for each instance can be retrieved by simply identifying the nearest prototype.

\subsection{Representation learning}
\label{sec:pcl}

The goal of representation learning is to learn a discriminative representation that can well separate different categories, not only the old ones, but also the new ones.
Contrastive learning (CL) has been shown to be an effective choice for NCD~\citep{jia2021joint} and GCD~\citep{vaze22generalized}. Self-supervised contrastive learning is defined as
\begin{equation}
    \mathcal{L}_{CL} = - \log \frac{\exp(z_i \cdot z_i' / \tau)}{\sum_{j=1}^n \exp(z_i \cdot z_j'/\tau)}
\end{equation}
where $z_i$ and $z_i'$ are the representations of two views obtained from the same image using random augmentations and $\tau$ is the temperature. Two views of the same instance are pulled closer, and different instances are pushed away during training. Self-supervised contrastive learning and its supervised variant, in which different instances from the same category are also pulled closer, are used in~\citet{vaze22generalized} for representation learning. However, as a stronger training signal is used for the labelled data, the representation is likely biased to the labelled data to some extent. Moreover, such a method does not take the downstream clustering task into account during learning, thus a clustering algorithm is required to run independently after the representation learning. 

In this paper, we adopt prototypical contrastive learning (PCL)~\citep{li2020prototypical} to the GCD setting to learn the representation $z_i=f(x_i) \in \mathbb{R}^d$.
PCL uses a set of prototypes $\mathcal{C}=\{\mu_1, \dots, \mu_K\}$ to represent the dataset for contrastive learning instead of the random augmentation generated views $z_i'$. PCL loss can be written as
\begin{equation}
    \mathcal{L}_{PCL} = - \log \frac{\exp(z_i \cdot \mu_s / \tau)}{\sum_{j=1}^K \exp(z_i \cdot \mu_j / \tau)}
    \label{eq:pcl}
\end{equation}
where $\mu_s$ is the corresponding prototype for $z_i$.
It was originally designed as an alternative for self-supervised contrastive learning by over-clustering the training data to obtain the prototypes during training. We employ PCL here to learn reliable representation while taking the downstream clustering into account for GCD, where we have a set of partially labelled data. In our case, the prototypes can be interpreted as the class centers for each of the categories.
To obtain the prototypes for the seen categories, we directly calculate the class mean by averaging all the feature vectors of the labelled instances. For the unseen categories, we obtain the prototypes with a semi-supervised variant of the Gaussian Mixture Model (GMM), as will be introduced in~\cref{sec:gmm}. This way, the cluster assignment for an unlabelled image can be readily achieved by finding the nearest prototype. 

Additionally, we observe that only a few principal dimensions can already recover most of the variances in the representation space of $z_i$, which is known as \textit{dimensional collapse} (DC) in~\citep{Jing2021UnderstandingDC,hua2021feature}, and it is shown that DC can be caused by strong augmentations or implicit regularizations in the model, and preventing DC during training can lead to a better feature representation.
To alleviate DC for representation learning in our case, we propose to first project the feature to a subspace obtained by principal component analysis (PCA) before the contrastive learning.
Specifically, we apply PCA on a matrix $Z$ formed by a mini-batch of features $z_i$, with a batch size of $n$, the feature dimension $d$, and the number of effective principal directions $q$. We have $Z \approx U diag(S) V^\top$,
where $U \in \mathbb{R}^{n\times q}$, $S \in \mathbb{R}^q$ and $V \in \mathbb{R}^{d\times q}$.
We can then project features $z_i$ to principal directions to obtain a more compact feature $v_i = V \cdot z_i$, and replace feature $z_i$ with $v_i$ in~\cref{eq:pcl} for PCL. The prototypes are also computed in the projected space.

We jointly use self-supervised contrastive learning and PCL to train our model. The overall learning objective can be written as
\begin{equation}
    \mathcal{L}= \mathcal{L}_{CL} + \lambda(t)\mathcal{L}_{PCL}
\end{equation}
where $\lambda(t)$ is a linear warmup function defined as $\lambda(t) = \min(1, \frac{t}{T})$ where $t$ is the current epoch and $T$ warmup length ($T=20$ in our experiments). 
The reason we use both CL and PCL is that, in the beginning, the representation is not well suited for clustering, and thus the obtained prototypes are not informative to facilitate the representation learning. Hence, we gradually increase the weight of PCL during training from 0 to 1 in the first $T$ epochs.

\begin{figure}[t]
    \centering
    \includegraphics[width=1\linewidth]{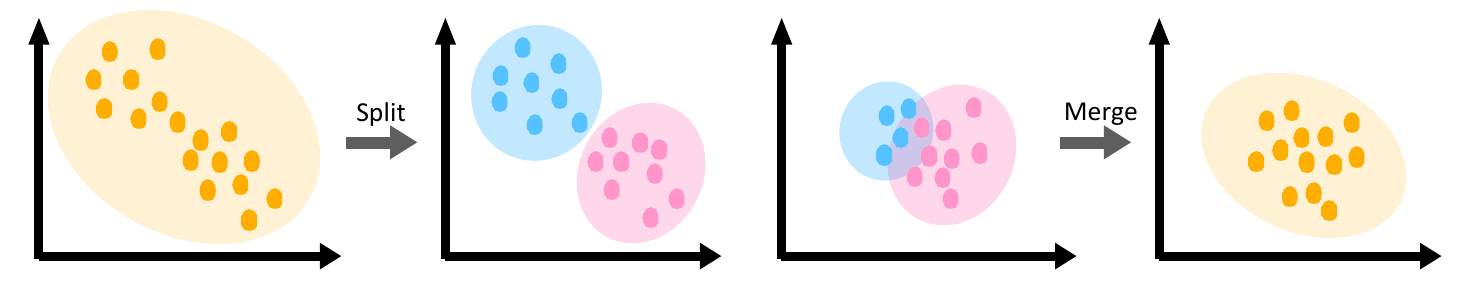}
    \caption{\textbf{Examples for splitting a separable cluster and merging two cluttered clusters.} Left: the cluster is split because the two sub-component in this cluster are easily separable. Right: two clusters are merged as they are cluttered and likely from the same class.}
    \label{fig:split_merge}
\end{figure}

\subsection{Class number and prototypes estimation with semi-supervised Gaussian mixture model}
\label{sec:gmm}


In this section, we present a semi-supervised variant of the Gaussian mixture model (GMM) with each Gaussian component consisting of two sub-components to estimate the prototypes for representation learning in~\cref{sec:pcl} and the unknown class number.
GMM estimates the prototypes and assigns a label for each data point by finding its nearest prototype. The cluster label assignment and the prototypes are then used for prototypical contrastive learning.
The GMM is defined as
\begin{equation}
    p(z) = \sum_{i=1}^K \pi_i \mathcal{N}(z|\mu_i, \Sigma_i),
\end{equation}
where $\mathcal{N}(z|\mu_i, \Sigma_i)$ is the Gaussian probability density function with mean $\mu_i \in \mathbb{R}^d$ and covariance $\Sigma_i \in \mathbb{R}^{d \times d}$, and $\pi_i$ is the weight for $i$-th Gaussian component and we have $\sum_{i=1}^N \pi_i =1$.
Ideally, we would expect the component number $K$ in the GMM to be equal to the class number $K^u$ in $\mathcal{D}$.
To estimate the unknown class number $K^u$, we leverage an automatic \textit{splitting-and-merging} strategy into the modeling process to obtain an optimal $K$, which is expected to be as close to $K^u$ as possible. We alternate between representation learning and $K^u$ estimation until convergence to get discriminative representation learning and a reliable class number estimation.
For initialization, $K$ can be set to any number greater than $K^l$. In our experiments, we simply set the initial number of components to a default $K_{init}=K^l+\frac{K^l}{2}$.
We run a semi-supervised $k$-means algorithm~\citep{vaze22generalized} with $k=K$ to obtain the $\mu$ and $\Sigma$ for each component in the mixture model.
Note that the semi-supervised $k$-means algorithm is constrained to the labelled data in a way that labelled instances from the same class are assigned to the same cluster, and labelled instances from different classes will not be assigned to the same cluster.
To facilitate the splitting and merging process, for each Gaussian component defined by $\mu_i$ and $\Sigma_i$, we further depict it with a GMM with two sub-components $\mu_{i, 1}, \mu_{i, 2}$ and $\Sigma_{i,1}, \Sigma_{i,2}$ with $\pi_{i,1}+\pi_{i,2}=1$.
We run a $k$-means with $k=2$ on the $i$-th component to obtain $\mu_{i, 1}, \mu_{i, 2}$ and $\Sigma_{i,1}, \Sigma_{i,2}$. 


\begin{algorithm}[t]
    \DontPrintSemicolon
    \SetAlgoLined
    \SetKwInOut{Input}{Inputs}
    \SetKwInOut{Output}{Output}
    \KwIn{
    \par
    \begin{tabular}{ll}
    $\mathcal{D}$, $\mathcal{D}^l$, and $\mathcal{D}^u$       &   The datasets. \\
    $K_{init}$                         & Initial guess of $K$. \\
    \end{tabular}}

    $K \gets K_{init}$ \\
    \For {$e=1$ \KwTo $E$}{
        $z\gets f(x), x\in\mathcal{D}$ \tcp*{extract features}
        $\mu, \Sigma \gets \arg\max \sum_{i=1}^K \pi_i \mathcal{N}(z|\mu_i, \Sigma_i)$  \tcp*{estimate prototypes using GMM}
        \For{$i=1$ \KwTo $len(\mathcal{D})$}{ 
           $\mathcal{B}^l \gets \{x_i^l\sim\mathcal{D}^l\}_{i=1}^{N^{l}}$ \tcp*{sample a batch of $N^l$ labelled images}
            $\mathcal{B}^u \gets \{x_i^u\sim\mathcal{D}^u\}_{i=1}^{N^{u}}$ \tcp*{sample a batch of $N^u$ unlabelled images}
            $f \gets \arg\min \mathcal{L}(f,\mu, \mathcal{B}^l, \mathcal{B}^u)$ \tcp*{prototypical contrastive learning}

        }
        $H_s, H_m \gets \text{calc\_prob}(\mu, \Sigma)$ \tcp*{probability for split and merge}
        $\mu, \Sigma \gets \text{perform\_op}(H_s, H_m)$ \tcp*{perform operations}
        $K \gets len(\mu)$ \tcp*{update $K$}
    }
    
    \KwOut{feature extractor $f(\cdot)$, cluster centers $\mu_i$}
    \caption{The overall algorithm of our method.}
    \label{alg:framework}
\end{algorithm}

For a cluster whose two sub-components are roughly independent and equally sized (\eg, left part of~\cref{fig:split_merge}), \ie, they are easily separable, we would like the model to split it into two such that the model can better fit the data distribution and the class assignment will be more accurate because it is less likely that such distinct clusters will belong to the same class.
For two clusters that are cluttered with each other (\eg, right part of~\cref{fig:split_merge}), \ie, difficult to distinguish, we would like to merge them into one, so that they will be considered as from the same class.
Following this intuition, we use the Metropolis-Hastings framework~\citep{hastings1970monte} to compute a probability $p_s = \min(1, H_s)$ to stochastically split a cluster into two. The Hastings ratio is defined as
\begin{align}
    H_{s}=\frac{\Gamma(N_{i,1})h(\mathcal{Z}_{i,1};\theta) \Gamma(N_{i,2})h(\mathcal{Z}_{i,2};\theta)}{\Gamma(N_i)h(\mathcal{Z}_i;\theta)},
\end{align}
where $\Gamma$ is the factorial function, \revision{\ie, $\Gamma(n)=n!=n\times(n-1)\times\dots\times1$}, $\mathcal{Z}_i$ is the set of data points in cluster $i$, $\mathcal{Z}_{i,j}$ is the set of data points in the $j$-th sub-cluster of cluster $i$, $N_i=|\mathcal{Z}_i|$, $N_{i,j}=|\mathcal{Z}_{i,j}|$, $h(Z;\theta)$ is the marginal likelihood of the observed data $\mathcal{Z}$ by integrating out the $\mu$ and $\Sigma$ parameters in the Gaussian, and $\theta$ is the prior distribution of $\mu$ and $\Sigma$. More details can be found in the supplementary. 
The intuition behind this $H_{s}$ is that, if the number of data points in two sub-components is roughly balanced, which is measured by the $\Gamma(\cdot)$ terms, and the data points in the two sub-components are independent of each other, which is measure by the $h(\cdot;\theta)$ terms, there should be a greater chance of splitting the cluster.
After performing a split operation, the $\mu_i$ and $\Sigma_i$ of previous components $i$ will be replaced with $\mu_{i,1}, \mu_{i,2}$ and $\Sigma_{i,1}, \Sigma_{i,2}$ of two sub-components.
We will then run two $k$-means within the two newly formed components to obtain their corresponding sub-components.
On the contrary, if two clusters are cluttered with each other, they should be merged. 
Similar to splitting, we determine the merging probability by $p_m = \min(1, H_m)$, where $H_m$ is calculated similarly for two clusters $i$ and $j$:
\begin{align}
    H_{m} = \frac{\Gamma(N_i + N_j) h(\mathcal{Z}_{i}\cup \mathcal{Z}_j; \theta)}{\Gamma(N_{i})h(\mathcal{Z}_{i};\theta) \Gamma(N_{j})h(\mathcal{Z}_{j};\theta)}.
\end{align}
Note that both $H_s$ and $H_m$ are within the range of $(0, +\infty)$, so we use $p_s=min(1, H_s)$ and $p_m=min(1, H_m)$ to convert it into a valid probability.

To take the labelled instances into consideration during the splitting-and-merging process, if a cluster consists of labelled instances, we set its $p_s = 0$; if for any two clusters containing instances from two labelled classes, we set their $p_m = 0$. 

During the splitting-and-merging process, we first apply splitting according to the $p_s$ and then apply merging according to $p_m$. The newly formed clusters by splitting will not be reused during the merging step. After finishing the splitting and merging, we can obtain the prototypes, and thus can estimate $K$, for our PCL-based representation learning. 
We alternate between representation learning and class number estimation for each training epoch until converge. The final $K$ will be considered the estimated class number in $\mathcal{D}$. The cluster assignment for each unlabelled instance can be easily retrieved by identifying its nearest prototype, without the need of running a non-parametric clustering algorithm as~\citet{vaze22generalized}. The overall training process is summarized in~\cref{alg:framework}.

%% file: cvpr/exp_cvpr.tex
\begin{table*}[t]
\footnotesize
\centering
\caption{\textbf{Results on generic image classification datasets.}}\label{tab:generic}
\resizebox{0.8\linewidth}{!}{ 
\begin{tabular}{clccccccccccc}
\toprule
&&   &     & \multicolumn{3}{c}{CIFAR10} & \multicolumn{3}{c}{CIFAR100} & \multicolumn{3}{c}{ImageNet-100}\\
\cmidrule(rl){5-7}
\cmidrule(rl){8-10}
\cmidrule(rl){11-13}
No. &Methods                                   & \makecell{Known\\$K$} & PCA & All  & Old  & New  & All  & Old  & New  & All  & Old  & New \\\midrule
(1) & $k$-means~\citep{macqueen1967some_kmeans} & \cmark & \xmark & 83.6 & 85.7 & 82.5 & 52.0 & \iclrrebuttal{52.2} & \iclrrebuttal{50.8} & \iclrrebuttal{72.7} & 75.5 & \uline{71.3} \\
(2) & RankStats+~\citep{han21autonovel}         & \cmark & \xmark & \iclrrebuttal{46.8} & \iclrrebuttal{19.2} & \iclrrebuttal{60.5} & \iclrrebuttal{58.2} & \iclrrebuttal{77.6} & \iclrrebuttal{19.3} & \iclrrebuttal{37.1} & \iclrrebuttal{61.6} & \iclrrebuttal{24.8} \\
(3) & UNO+~\citep{fini2021unified}              & \cmark & \xmark & \iclrrebuttal{68.6} & \uline{\iclrrebuttal{98.3}} & \iclrrebuttal{53.8} & \iclrrebuttal{69.5} & \iclrrebuttal{80.6} & \iclrrebuttal{47.2} & \iclrrebuttal{70.3} & \uline{\iclrrebuttal{95.0}} & \iclrrebuttal{57.9} \\
(4) & ORCA~\citep{cao21orca}                     & \cmark & \xmark & \iclrrebuttal{81.8} & \iclrrebuttal{86.2} & \iclrrebuttal{79.6} & \iclrrebuttal{69.0} & \iclrrebuttal{77.4} & \iclrrebuttal{52.0} & \iclrrebuttal{73.5} & \iclrrebuttal{92.6} & \iclrrebuttal{63.9} \\
(5) & Vaze \etal~\citet{vaze22generalized}                 & \cmark & \xmark & 91.5 & 97.9 & 88.2 & \iclrrebuttal{73.0} & \iclrrebuttal{76.2} & \uline{\iclrrebuttal{66.5}} & \iclrrebuttal{74.1} & \iclrrebuttal{89.8} & \iclrrebuttal{66.3} \\
\midrule
(6) & \textit{Ours} (GPC)                       & \cmark & \xmark & 92.0 & \uline{98.3} & 88.7 & 77.4 & 84.8 & 62.4 & 76.5 & 94.0 & 68.5 \\ 
(7) & \textit{Ours} (GPC)                       & \cmark & \cmark & \uline{92.2} & 98.2 & \uline{89.1} & \uline{77.9} & \uline{85.0} & 63.0 & \uline{76.9} & 94.3 & 71.0 \\
\midrule
\rowcolor{shaded_gray}(8) & Vaze \etal~\citet{vaze22generalized}                 & \xmark & \xmark & 88.6 & 96.2 & 84.9 & 73.2 & 83.5 & 57.9 & 72.7 & 91.8 & 63.8 \\
\rowcolor{shaded_gray}(9) & Vaze \etal~\citet{vaze22generalized}                 & \xmark & \cmark & 89.7 & 97.3 & 86.3 & 74.8 & 83.8 & 58.7 & 73.8 & 92.1 & 64.6 \\
\midrule
\rowcolor{shaded_gray}(10) & \textit{Ours} (GPC)                       & \xmark & \xmark & 88.2 & 97.0 & 85.8 & 74.9 & 84.3 & 59.6 & 74.7 & 92.9 & 65.1 \\
\rowcolor{shaded_gray}(11) & \textit{Ours} (GPC)                      & \xmark & \cmark & \textbf{90.6} & \textbf{97.6} & \textbf{87.0} & \textbf{75.4} & \textbf{84.6} & \textbf{60.1} & \textbf{75.3} & \textbf{93.4} & \textbf{66.7} \\
\bottomrule
\end{tabular}
}
\end{table*}

\section{Experimental results}
\label{sec:exp}

\subsection{Experimental setup}
\label{sec:exp_setup}

\paragraph{Benchmark and evaluation metric.}
We validate the effectiveness of our method on the generic image classification benchmark (including CIFAR-10/100~\citep{cifar} and ImageNet-100~\citep{cmc}) and also the recently proposed Semantic Shift Benchmark~\citep{vaze22openset} (SSB)(including CUB-200~\citep{cub200}, Stanford Cars~\citep{stanfordcars}, and FGVC-Aircraft~\citep{aircraft}).
For each of the datasets, we follow~\citet{vaze22generalized} and sample a subset of all classes for which we have annotated labels during training. 
For experiments on SSB datasets, we directly use the class split from~\citet{vaze22openset}.
50\% of the images from these labelled classes will be used as the labelled instances in $\mathcal{D}^l$, and the remaining images are regarded as the unlabelled data $\mathcal{D}^u$ containing instances from labelled and unlabelled classes.
See~\cref{tab:datasplit} for statistics of the datasets we evaluated.
We evaluate model performance with clustering accuracy (ACC) following standard practice in the literature. 
At test-time, given ground truth labels $y^*$ and model predicted cluster assignments $\hat{y}$, the ACC is calculated as $ACC = \frac{1}{M} \sum_{i=1}^{M} \mathds{1}(y^*_i = g(\hat{y}_i))$ where $g$ is the optimal permutation for matching predicted cluster assignment $\hat{y}$ to actual class label $y^*_i$ and $M = |\mathcal{D}^u|$.

\begin{table}[h]
    \centering
    \caption{\textbf{Data splits in the experiments.}}\label{tab:datasplit}
    \begin{tabular}{lcc}
    \toprule
    & labelled  & unlabelled  \\
    \midrule
    CIFAR-10 & 5 & 5 \\
    CIFAR-100 & 80 &  20 \\
    ImageNet-100 & 50 & 50 \\
    \midrule
    CUB-200 & 100 & 100 \\
    Stanford-Cars & 98 &  98 \\    
    FGVC-aircraft & 50 &  50 \\   
    \bottomrule
    \end{tabular}
    \vspace{-1.5em}
\end{table}

\paragraph{Implementation details.}
We train and test all the methods with a ViT-B/16 backbone~\citep{dosovitskiy2020vit} with pretrained weights from DINO~\citep{caron2021emerging}. We use the output of \texttt{[CLS]} token with a dimension of 768 as the feature representation for an input image.
We only finetune the last block of the ViT-B backbone to prevent the model from overfitting to the labelled classes during training.
We set the batch size for training the model to 128 with 64 labelled images and 64 unlabelled images and use a cosine annealing schedule for the learning rate starting from 0.1.
The number of principal directions in the PCA is set to 128, which we found performs the best across all the datasets evaluated.
We train all the methods for 200 epochs on each dataset for a fair comparison with previous works, and the best-performing model is selected using the accuracy on the validation set of the labelled classes.
All experiments are done with an NVIDIA V100 GPU with 32GB memory.


\begin{table*}[t]
\footnotesize
\centering
\caption{\textbf{Results on Semantic Shift Benchmark datasets.}}\label{tab:fine_grained}
\resizebox{0.8\linewidth}{!}{ 
\begin{tabular}{clccccccccccc}
\toprule
&&&& \multicolumn{3}{c}{CUB} & \multicolumn{3}{c}{Stanford Cars} & \multicolumn{3}{c}{FGVC-aircraft}\\
\cmidrule(rl){5-7}
\cmidrule(rl){8-10}
\cmidrule(rl){11-13}
No. & Methods                                  & \makecell{Known\\$K$} & PCA & All  & Old  & New  & All  & Old  & New  & All  & Old  & New \\
\midrule
(1) & $k$-means~\citep{macqueen1967some_kmeans} & \cmark & \xmark & 34.3 & 38.9 & 32.1 & 12.8 & 10.6 & 13.8 & 16.0 & 14.4 & 16.8 \\
(2) & RankStats+~\citep{han21autonovel}         & \cmark & \xmark & 33.3 & 51.6 & 24.2 & 28.3 & 61.8 & 12.1 & 26.9 & 36.4 & 22.2 \\
(3) & UNO+~\citep{fini2021unified}              & \cmark & \xmark & 35.1 & 49.0 & 28.1 & 35.5 & \uline{70.5} & 18.6 & 40.3 & \uline{56.4} & 32.2 \\
(4) & ORCA~\citep{cao21orca}                     & \cmark & \xmark & \iclrrebuttal{35.3} & \iclrrebuttal{45.6} & \iclrrebuttal{30.2} & \iclrrebuttal{23.5} & \iclrrebuttal{50.1} & \iclrrebuttal{10.7} & \iclrrebuttal{22.0} & \iclrrebuttal{31.8} & \iclrrebuttal{17.1} \\
(5) & Vaze \etal~\citet{vaze22generalized}                 & \cmark & \xmark & 51.3 & 56.6 & 48.7 & 39.0 & 57.6 & 29.9 & 45.0 & 41.1 & 46.9 \\
\midrule
(6) & \textit{Ours} (GPC)                      & \cmark & \xmark & 54.2 & 54.9 & 50.3 & 41.2 & 58.8 & 31.6 & 46.1 & 42.4 & 47.2 \\ 
(7) & \textit{Ours} (GPC)                      & \cmark & \cmark & \uline{55.4} & \uline{58.2} & \uline{53.1} & \uline{42.8} & 59.2 & \uline{32.8} & \uline{46.3} & 42.5 & \uline{47.9} \\
\midrule
\rowcolor{shaded_gray}(8) & Vaze \etal~\citet{vaze22generalized}      & \xmark & \xmark & 47.1 & 55.1 & 44.8 & 35.0 & 56.0 & 24.8 & 40.1 & 40.8 & 42.8 \\
\rowcolor{shaded_gray}(9) & Vaze \etal~\citet{vaze22generalized}      & \xmark & \cmark & 49.2 & \textbf{56.2} & 46.3 & 36.3 & 56.6 & 25.9 & 43.2 & \textbf{40.9} & 44.6 \\
\midrule
\rowcolor{shaded_gray}(10) & \textit{Ours} (GPC)                     & \xmark & \xmark & 50.2 & 52.8 & 45.6 & 36.7 & 56.3 & 26.3 & 39.7 & 39.6 & 42.7 \\
\rowcolor{shaded_gray}(11) &\textit{Ours} (GPC)                     & \xmark & \cmark & \textbf{52.0} & 55.5 & \textbf{47.5} & \textbf{38.2} & \textbf{58.9} & \textbf{27.4} & \textbf{43.3} & 40.7 & \textbf{44.8} \\
\bottomrule
\end{tabular}
}
\end{table*}

\subsection{Comparison with the state-of-the-art}


In~\cref{tab:generic}, we report the comparison with the state-of-the-art method of \citet{vaze22generalized}, strong baselines derived from NCD methods, and the $k$-means on the generic classification datasets. Notably, our method consistently achieves the best overall performance on all datasets, under the challenging setting where the class number is unknown.
When the class number is known, our method also achieves the best performance on all datasets, except ImageNet-100, on which the best performance is achieved by ORCA~\citep{cao21orca}.
In rows 1-7, we compare with other methods with the known class number in the unlabelled data, while in rows 8-11 we compare with \citet{vaze22generalized} for the case of the unknown class number.
We can see that our proposed framework outperforms other methods in most cases and especially when the number of classes is unknown.
Comparing rows 10 and 11 to row 5, we can see that our proposed method without knowing the number of classes can even match the performance of previous strong baseline with the number of classes known to the model.
Furthermore, from row 6 vs row 7 and row 10 vs row 11,  we can see that the additional PCA layer can effectively improve the performance, also the performance improvement from PCA is larger on the `New' classes than on the `Old' classes, which validates that the PCA can keep the representation space from collapsing and improve the performance on classes without using any labels.
Due to the fact that labelled instances provide a stronger training signal, we can see from rows 6 - 11 that performance on `Old' classes is generally steady.
\cvprrevision{
Comparing row 7 to row 5 and row 11 to row 8,  we can see our full method outperforms the previous state-of-the-art method Vaze~\etal~\cite{vaze22generalized} by large margins on both known and unknown class number cases}.
\cref{tab:fine_grained} shows performance comparison on the more challenging fine-grained Semantic Shift Benchmark~\citep{vaze22openset}. A similar trend of~\cref{tab:generic} holds true for the results on SSB.
Our approach achieves competitive performance in all cases and again reaches a better performance when the number of classes is unknown.
In~\cref{tab:herb}, we present the results on the Herbarium-19 dataset which is a long-tailed dataset, adding additional challenges for the GCD task. Again, our method performs the best on `All' and `New' classes.
These results demonstrate the effectiveness of our method.

\begin{table}[t]
\centering
\caption{Results on the Herbarium 19 Dataset}
\label{tab:herb}
\begin{tabular}{lccc}
\toprule
Methods & All & Old & New \\
\midrule
$k$-means~\cite{macqueen1967some_kmeans} & 13.0 & 12.2 & 13.4 \\
RS+~\cite{han21autonovel} & 27.9 & \textbf{55.8} & 12.8 \\
UNO+~\cite{fini2021unified} & 28.3 & \uline{53.7} & 14.7 \\
ORCA~\cite{cao21orca} & 20.9 & 30.9 & 15.5 \\
GCD~\cite{vaze22generalized} & \uline{35.4} & 51.0 & \uline{27.0} \\
\midrule
Ours (GPC) & \textbf{36.5} & 51.7 & \textbf{27.9} \\
\bottomrule
\end{tabular}
\end{table}

\begin{figure}
\centering
\begin{subfigure}{.4\textwidth}
  \centering
  \includegraphics[width=\linewidth]{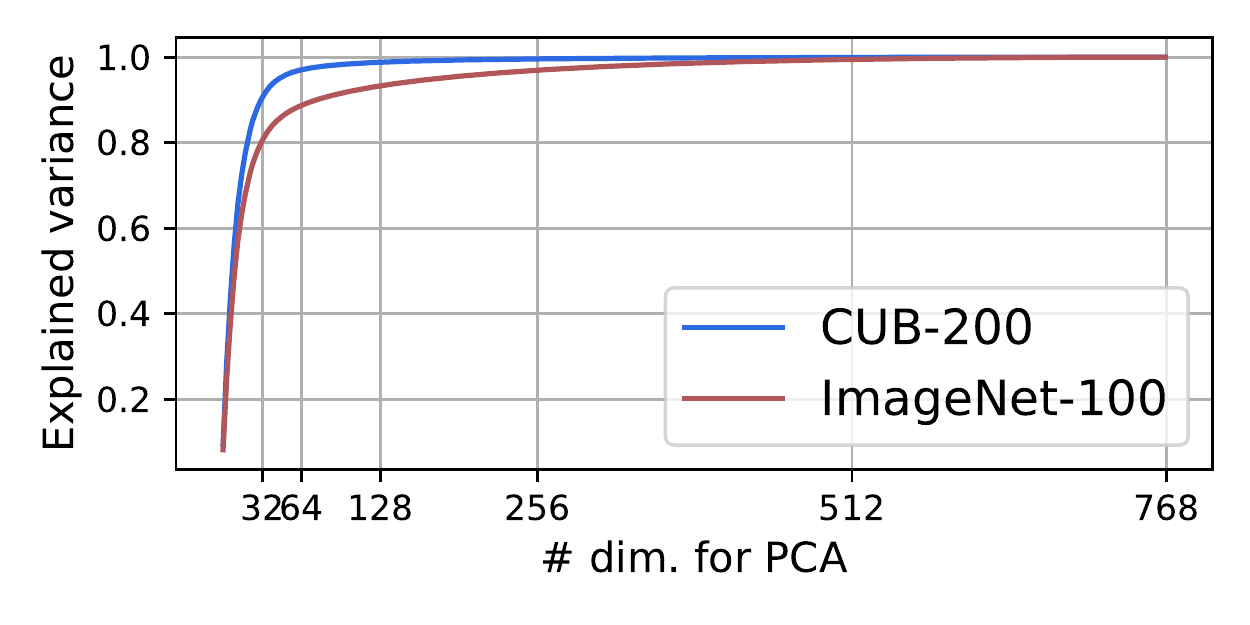}
  \caption{Explained variance of the original feature \wrt\ the number of principal directions.}
  \label{fig:pca_var}
\end{subfigure}%
\hspace{2em}
\begin{subfigure}{.4\textwidth}
  \centering
  \includegraphics[width=\linewidth]{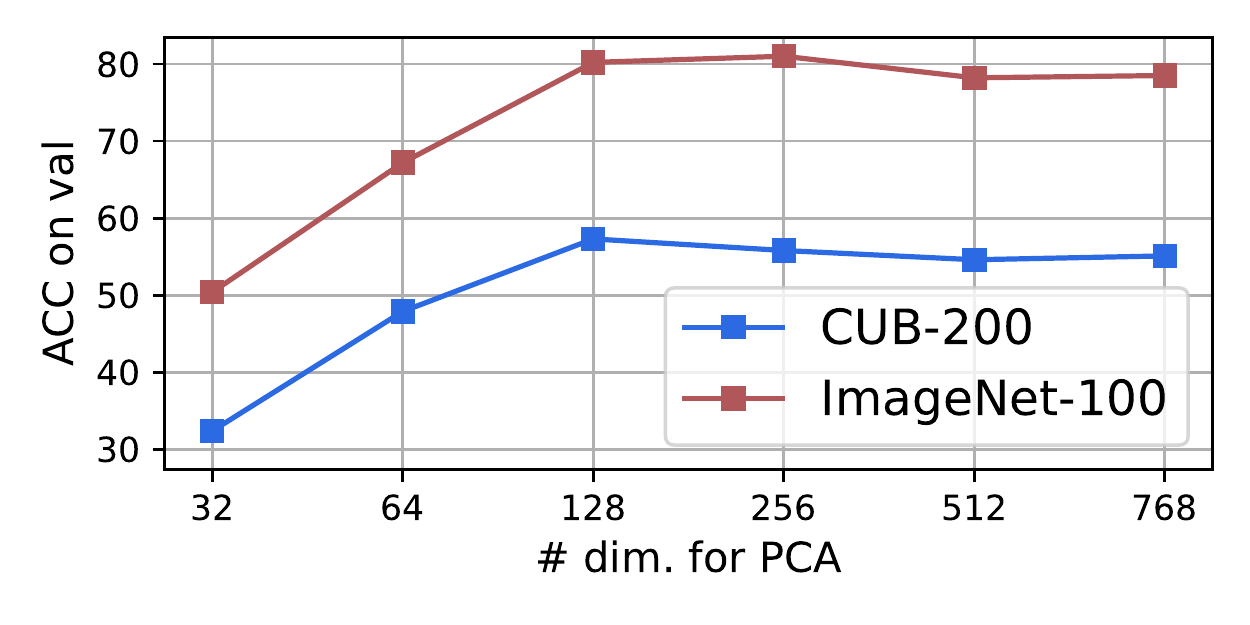}
  \caption{The clustering ACC on validation set \wrt\ the number of principal directions}
  \label{fig:pca_acc}
\end{subfigure}
\caption{\textbf{The effects of the number of principal directions in PCA on the feature representations.}}
\label{fig:pca}
\vspace{-4pt}
\end{figure}


\subsection{Novel class number estimation}

One of the important yet overlooked components in the NCD and GCD literature is the estimation of unknown class numbers.
Our proposed framework leverages a modified GMM to estimate the class number, in which we need to define an initial guess of the class number. We validate the effects of different choices of the initial guess $K_{init}$ \wrt\ the estimated class number in~\cref{tab:vary_k_init}. Note that the number in~\cref{tab:vary_k_init} is $K^{n}_{init} = K_{init}-K^l$. We can see that our proposed framework is generally robust to a wide range of initial guesses. We found that $K_{init}=K^l+\frac{K^l}{2}$ is a simple and reliable choice. Hence we use this for all datasets.

\begin{table*}[htb]
\centering
\caption{\textbf{Results of varying the initial guessed $K^n_{init}$.} `GT $K^{n}$' is the ground truth number of novel classes. $K^n$ is the estimated number of novel classes. Our proposed framework is generally robust in estimating the number of novel classes, and we found that using the initial guess of $K_{init}=K^l+\frac{K^l}{2}$ can be a simple and reliable choice.}
\label{tab:vary_k_init}
\setlength{\tabcolsep}{5pt}
\begin{tabular}{lcc|c|rcccccc}
\toprule 
Dataset          &  $K^{l}$  & GT $K^{n}$   & \makecell{Vaze \etal \citet{vaze22generalized}} &   $K^{n}_{init}=$ \,\,\,3  & 5      & 10 & 20 & 30   & 50    & 100   \\
\midrule
CIFAR-10         &  5        & 5            & 4      &  $K^{n}=$\,\,\, 5     &    5   & 5  & 6  &  6   & 8     & 14   \\ 
CIFAR-100        &  80       & 20           & 20     &  $K^{n}=$   16    &   20   & 20 & 21 &  22 & 27    &  36  \\
ImageNet-100     &  50       & 50           & 59     &  $K^{n}=$   58    &   48   & 57 & 55 & 54  & 50    & 60   \\
\midrule
CUB              &  100      &  100         & 131    &  $K^{n}=$   79    &   87   & 86 & 88 & 92  &  112  &  101  \\
SCars            &  98       &   98         & 132    &  $K^{n}=$   84    &   90   & 86 & 87 & 89  &  115  &  104  \\
\bottomrule
\end{tabular}
\vspace{-3pt}
\end{table*}

\subsection{Training complexity}
Given that our framework necessitates the fitting of GMMs, an extended training duration is required.
We compare the performance of our method to the extended method of Vaze~\etal~\cite{vaze22generalized}, which is pushed to $1.5\times$ its original number of training epochs.
The results in~\cref{tab:complex} demonstrate that our method outperforms ~\cite{vaze22generalized} across both CUB and ImageNet-100 datasets while maintaining a similar training duration.
Furthermore, in contrast to the baseline method, our approach eliminates the need for additional post-training procedures such as running SS-$k$-means on the entire dataset for label assignment.

\begin{table}[h]
\setlength\tabcolsep{3pt}
\centering
\caption{Performance Comparison of Vaze etal (1.5x) and GPC on CUB and ImageNet-100 Datasets}
\label{tab:complex}
\begin{tabular}{lcccccc}
\toprule
Method & \multicolumn{3}{c}{CUB} & \multicolumn{3}{c}{ImageNet-100} \\
\cmidrule(rl){2-4}
\cmidrule(rl){4-7}
& All & Old & New & All & Old & New \\
\midrule
Vaze \etal~\cite{vaze22generalized}        & 51.3 & 56.6 & 48.7 & 74.1 & 89.8 & 66.3 \\
Vaze \etal (1.5$\times$) & 52.0 & 56.8 & 49.0 & 74.7 & 90.3 & 66.7 \\
GPC & 55.4 & 58.2 & 53.1 & 76.9 & 94.3 & 71.0 \\
\bottomrule
\end{tabular}
\end{table}

\subsection{Ablation study}


\paragraph{Number of dimensions in PCA}

The PCA in our framework requires setting a number for the number of principal directions to extract from data.
In~\cref{fig:pca}, we show the results of using a different number of principal directions in PCA on CUB-200 and ImageNet-100 datasets.
We can see from~\cref{fig:pca_var} that for both datasets, 128 principal directions can already explain most of the variances in the data, thus we choose the PCA dimension to be 128 for all our experiments. We further experiment with other different choices of the PCA dimension and shows the result in~\cref{fig:pca_acc}, which again confirms that 128 principal directions are already expensive enough, and obtain the best performance over other choices, that are either too few or too many, effectively avoiding DC.

\paragraph{Different methods for prototype estimation}

Our semi-supervised GMM plays an important role in prototype estimation for representation learning based on prototypical contrastive learning. 
Here, we replace our semi-supervised GMM with other alternatives that do not produce prototypes automatically.
Particularly, we compare our method with DBSCAN~\citep{ester1996density}, Agglomerative clustering~\citep{murtagh2014ward}, and semi-supervised $k$-means~\citep{Han2019learning,vaze22generalized}. 
The prototypes are then obtained by averaging the data points that are assigned to the same cluster. For a fair comparison, the same regulations  to prevent the wrong clustering results for labelled instances are applied to all methods, \ie, during the clustering process, two labelled instances with the same label will fall into the same cluster, and two instances with different labels will be assigned to different clusters. The results are reported in~\cref{tab:pcl_with_other_cluster}.
Our method achieves the best performance on all three datasets, indicating that better prototypes are obtained by our approach to facilitate representation learning. Note that DBSCAN requires two important user-defined parameters, radius, and minimum core points, the ideal values of which lack a principled way to obtain in practice, while our method is parameter-free and can seamlessly be combined with the representation learning to jointly enhance each other, obtaining better performance.  

\begin{table}[t]
    \centering
    \caption{\textbf{Combining components of GPC with other methods.}
    ``IM-100'' denotes ImageNet-100.}
    \label{tab:ablation_combination}
    \begin{subtable}[h]{0.5\textwidth}
    \centering
    \caption{Different prototype estimation methods.}
    \label{tab:pcl_with_other_cluster}
        \begin{tabular}{c|ccc}
        \toprule
        Clustering Algo.                   & CUB  & IM-100 & SCars  \\
        \midrule
        Ester \etal~\citet{ester1996density}           & 45.6 &  66.1     & 34.8  \\
        Murtagh \etal~\citet{murtagh2014ward}            & 52.1 &  74.6     & 39.8  \\
        Vaze \etal~\citet{vaze22generalized}          & 49.2 &  73.2     & 37.4  \\
        \midrule                         
        \textit{Ours} (GPC)                & 54.1  & 76.6      &  41.9   \\
        \bottomrule
        \end{tabular}
        \vspace{1em}
    \end{subtable}
    \centering
    \begin{subtable}[h]{0.5\textwidth}
        \centering
        \caption{Combining our GMM with other methods.}
        \label{tab:gmm_with_other_rep}
        \begin{tabular}{c|ccc}
        \toprule
        Representation                     & CUB  & IM-100      & SCars \\
        \midrule                                             
        Han \etal~\citet{han21autonovel}             & 34.6 &  38.4       & 29.3 \\
        Zhao \etal~\citet{zhao21novel}                & 37.8 &  39.7       & 33.2 \\
        Vaze \etal~\citet{vaze22generalized}           & 50.6 &  73.4       & 37.8 \\
        \midrule                         
        \textit{Ours} (GPC)                & 54.1 & 76.6      &  41.9   \\
        \bottomrule
        \end{tabular}
     \end{subtable}
     \vspace{-7pt}
\end{table}

\paragraph{Combining our GMM with other GCD methods} 
We further combine our semi-supervised GMM with automatic splitting and merging with other methods, allowing joint representation learning and category discovery without a predefined category number. 
As the state-of-the-art GCD method~\citep{vaze22generalized} does not contain any parametric classifier during representation, so it can be directly combined with our GMM.
For the RankStat and the DualRank methods that have a parametric classifier for category discovery, we treat the weights of the classifier as the cluster centers and run our GMM to automatically determine the category number during representation learning. The results are presented in~\cref{tab:gmm_with_other_rep}. Comparing with row 9 in~\cref{tab:generic} and~\cref{tab:fine_grained}, we can see using our GMM can also improve~\citet{vaze22generalized} on CUB and Stanford Cars, while our proposed framework consistently achieves better performance on all datasets, again validating that our design choices.

\paragraph{Class number estimation with different representations}
Here, we validate our class number estimation method on top of the representations learned by other GCD approaches and report the results in~\cref{tab:estimated_k_other_rep}.
It can be seen, applying our method on other GCD representations can achieve reasonably well results. Notably, by applying our class number estimation method on top of the representation by the existing state-of-the-art method, we can obtain better class number estimation results, though the overall best results are obtained with the representation learned in our framework.

\begin{table}[t]
        \centering
        \caption{
        \textbf{Class number estimation with different learned representations.} ``C-100'' stands for the CIFAR-100 dataset.}
        \label{tab:estimated_k_other_rep}
        \begin{tabular}{c|ccccc}
        \toprule
        Representation                      &  C-100 & CUB & SCars & IM-100  \\
        \midrule
        Ground Truth $K^n$                  &    20      & 100 & 98    & 50 \\
        \midrule
        Vaze~\etal\citet{vaze22generalized}           &     20    &   131  &  132       &  59      \\
        \midrule
        \textit{Ours} w/ \citet{han21autonovel} feat.         &   19      & 111 & 94    &  55     \\
        \textit{Ours} w/ \citet{zhao21novel} feat.            &   22      & 116 & 89    &  49     \\
        \textit{Ours} w/ \citet{vaze22generalized} feat.      &   21      & 121 & 109   &  57        \\
        \midrule                         
        \textit{Ours} (GPC)                        &   20      & 112 & 103   & 53      \\
        \bottomrule
        \end{tabular}
\end{table}

\paragraph{Partial overlap between $\mathcal{Y}_l$ and $\mathcal{Y}_u$}
We evaluate the performance of our method when we relax $\mathcal{Y}_l \subset \mathcal{Y}_u$ to $\mathcal{Y}_l\cap\mathcal{Y}_u\neq \emptyset$, \ie, the two sets may only partially overlap. We vary the number of overlapped classes and report the results in~\cref{tab:partial}.

Our approach consistently outperforms the method proposed by Vaze~\etal~\cite{vaze22generalized} in all configurations. These compelling results not only showcase the robustness of our method but also highlight its effectiveness in scenarios involving partial overlap between known and unknown classes.

\begin{table}[h]
     \setlength\tabcolsep{3pt}
    \centering
    \caption{Results on CUB of only partial overlap between $\mathcal{Y}_l$ and $\mathcal{Y}_u$.}
    \label{tab:partial}
    \footnotesize
    \begin{tabular}{c ccc ccc ccc} 
    \toprule
    $|\mathcal{Y}_l\cap\mathcal{Y}_u|$     & \multicolumn{3}{c}{25} & \multicolumn{3}{c}{50} & \multicolumn{3}{c}{75} \\
    \cmidrule(rl){2-4}\cmidrule(rl){5-7}\cmidrule(rl){8-10}
    & All & Old & New & All & Old & New &  All & Old & New \\
    \midrule
    Vaze~\etal~\cite{vaze22generalized}                             & 49.5 & 50.1 & 48.2  & 51.2 & 50.7 & 52.2 & 52.7 & 50.9 & 54.5 \\
    \rowcolor{baselinecolor}Ours           & 51.2 & 52.6 & 49.5  & 52.3 & 51.6 & 54.8 & 53.6 & 51.4 & 55.9 \\
    \bottomrule
    \end{tabular}
\end{table}

\paragraph{Varying ratio of Old/New categories}

We measure the estimated new class numbers when varying the ratio of Old/New categories while having $K_{init}=K^l+\frac{K^l}{2}$ on CUB in~\cref{tab:vary}. As can be seen, for all cases, our method outperforms~\cite{vaze22generalized}. Meanwhile, we can also see that when the initial guess is too far from the ground truth, the estimation will be less accurate.


\begin{table}[ht]
    \centering
    \caption{Estimated class numbers on CUB with a varying ratio of Old/New classes}
    \label{tab:vary}
    \begin{tabular}{ccccc}
    \toprule
    Old/New     & 20/180   & 40/160  & 60/140 & 80/120 \\
    \midrule
    GCD~\cite{vaze22generalized} &  87   &   102   &  114  &  104    \\
    Ours        &  93     &  126    &  135   &  116   \\
    \bottomrule
    \end{tabular}
\end{table}


%% file: cvpr/conclusion_cvpr.tex
\section{Conclusion}
\label{sec:conclusion}
In this paper, we present an EM-like framework for the challenging GCD problem without knowing the number of new classes, with the E-step automatically determining the class number and prototypes and the M-step being robust representation learning. We introduce a semi-supervised variant of GMM with a stochastic splitting and merging mechanism to obtain the prototypes and leverage these evolving prototypes for representation learning by prototypical contrastive learning. We demonstrated that class number estimation and representation learning can facilitate each other for more robust category discovery. Our framework obtains state-of-the-art performance on multiple public benchmarks. 


\paragraph{Acknowledgements}
This work is supported by
Hong Kong Research Grant Council - Early Career Scheme (Grant No. 27208022) and HKU Seed Fund for Basic Research.

%% file: cvpr/supp_cvpr.tex
\newpage





\section{Details of the framework for splitting and merging clusters}

In this section, we provide details for the Metropolis-Hastings framework.
In the Gaussian Mixture Model, we have three sets of parameters, $(\pi_i, \mu_i, \Sigma_i)$ where $\pi_i$ is the mixture weight and $\mu_i, \Sigma_i$ are the mean and covariance matrix. 
These parameters are assumed to be sampled from a prior distribution.
When $\mu_i$ and $\Sigma_i$ are unknown for the multivariate Gaussian distribution, we adopt the Normal Inverse Wishart (NIW) distribution as the prior to sample them for algebraic convenience, because NIW distribution is a conjugate prior and the conjugacy property can lead to a closed-form expression of the posterior.

The Inverse Wishart (IW) distribution is defined as follows:
\begin{equation}
    p(\Sigma_i)\sim\mathcal{W}^{-1}(\nu, \Psi)=\frac{|\nu \Psi|^{\frac{\nu}{2}}}{2^{\frac{\nu d}{2}}\Gamma_d(\frac{\nu}{2})} |\Sigma_i|^{-\frac{\nu + d+1}{2}} \exp(-\frac{1}{2}tr(\nu \Psi \Sigma_i^{-1})),
\end{equation}
where $\Sigma_i$ is a $d\times d$ Symmetric and Positive Definite(SPD) matrix, $\nu > d-1$, $\Psi \in \mathbbm{R}^{d\times d}$ is SPD, and $\Gamma_d$ is a $d$-dimensional multivariate factorial function. The positive real number $\nu$ and the SPD matrix $\Psi$ are the parameters of the IW distribution. 
The data distribution determined by $\mu_i$ and  $\Sigma_i$ follows NIW distribution, if the joint probability density function is defined by
\begin{equation}
    p(\mu_i,\Sigma_i)\sim \text{NIW}(\kappa, \mathbf{m}, \nu, \Psi)\triangleq \mathcal{N}(\mu_i;\mathbf{m},\frac{1}{\kappa} \Sigma_i) \mathcal{W}^{-1}(\Sigma_i;\nu,\Psi),
\end{equation}
where $\mathbf{m}\in \mathbbm{R}^d$, $\kappa > 0$, and $\mathcal{N}(\mu_i; \mathbf{m}, \frac{1}{\kappa} \Sigma_i)$ is a $d$-dimensional Gaussian with mean $\mathbf{m}$ and covariance $\frac{1}{\kappa} \Sigma_i$ evaluated at $\mu_i$.

Given a set of features $\mathcal{Z}_i$ (with $N_i = |\mathcal{Z}_i|$) assigned to the Gaussian component $\mu_i, \Sigma_i$, we can have a posterior distribution of $\mu_i, \Sigma_i$ in a closed-form thanks to the conjugacy:
\begin{equation}
    p(\mu_i, \Sigma_i| \mathcal{Z}_i)=\text{NIW}(\mu_i, \Sigma_i; \kappa^{*}, \mathbf{m}_i^{*}, \nu^{*}, \Psi_{i}^{*}),
\end{equation}
where the posterior parameters are obtained by:
\begin{align}
    \kappa_{i}^* &= \kappa + N_i \\
    \mathbf{m}_i^* &= \frac{1}{\kappa_i^*}  [\kappa \mathbf{m} + \sum_{z_k\in \mathcal{Z}} z_k]\\
    \nu_i^* &= \nu + N_i \\
    \Psi_i^* &= \frac{1}{\nu^*} [\nu \Psi + \kappa \mathbf{m} \mathbf{m}^\top + (\sum_{z_k \in \mathcal{Z}_i} z_k z_k^\top) - \kappa_i^* \mathbf{m}_i^* \mathbf{m}_i^{*^\top}]
\end{align}

In Eq.5 and 6 of the main paper, we need to calculate the marginal likelihood function of the observed data $\mathcal{Z}_i$ by integrating out the $\mu_i$ and $\Sigma_i$ parameters in the Gaussian. Let $\theta = (\mathbf{m}, \kappa, \Psi, \nu)$ be the parameters of the NIW distribution. The marginal likelihood can be defined as follows:
\begin{align}
    h(\mathcal{Z}_i;\theta)&=\int p(\mathcal{Z}_i|\mu_i, \Sigma_i) p(\mu_i, \Sigma_i;\theta) d(\mu_i, \Sigma_i) \\
&=\frac{1}{\pi^{\nicefrac{Nd}{2}}} \frac{\Gamma_d (\nicefrac{\nu^*}{2})}{\Gamma_d (\nicefrac{\nu}{2})} \frac{|\nu \Psi|^{\nicefrac{\nu}{2}}}{|\nu^* \Psi_i^*|^{\nicefrac{\nu^*}{2}}} \frac{\kappa ^{\nicefrac{d}{2}}}{\kappa^{*^{\nicefrac{d}{2}}}},
\end{align}
with which we can compute the Eq. 5 and 6 in the main paper.



\section{Estimating the number of clusters on validation set}
In this section, we validate the choice of $K^n_{init}$ using only the labelled data, to better reflect the real world use case. In particular, we further split the classes in the labelled data $\mathcal{D}^l$ into two parts, $\mathcal{D}^l_r$ and $\mathcal{D}^l_p$. We drop the labels in $\mathcal{D}^l_p$. We verify the effectiveness of different choice of $K^n_{init}$ on $\mathcal{D}^l_p$ and report the results in~\cref{tab:vary_k_init_val}.
Interestingly, we observe that the initial guess of a number around $\frac{K^l}{2}$ often leads to a good estimate.

\begin{table}[htb]
\centering
\caption{\textbf{Results of varying the initial guessed $K^n_{init}$.} `GT $K^{n}$' is the ground truth number of novel classes, splited from the labelled set. $K^n$ is the estimated number of novel classes.}
\label{tab:vary_k_init_val}
\setlength{\tabcolsep}{5pt}
\begin{tabular}{lccrcccccc}
\toprule 
Dataset          &  $K^{l}$  & GT $K^{n}$ &$K^{n}_{init}=$ \,\,\,1 & 3      & 5  & 10 & 20   & 25    & 50  \\
\midrule
CIFAR-10         &  3        & 2            &  $K^{n}=$\,\,\, 2    &    2   & 4  & 5  &  3   & 6     & 8   \\
CIFAR-100        &  60       & 20           &  $K^{n}=$      15    &   18   & 18 & 20 & 21   & 22    & 29  \\
ImageNet-100     &  25       & 25           &  $K^{n}=$      18    &   19   & 22 & 21 & 23   & 27    & 29  \\
\midrule
CUB              &  50      &  50           &  $K^{n}=$      38    &   37   & 41 & 46 & 49   &  52  &  50  \\
SCars            &  49      &  49           &  $K^{n}=$      39    &   38   & 40 & 42 & 43   &  48  &  51  \\
\bottomrule
\end{tabular}
\end{table}

\iccvrevision{


\section{Convergence of the estimated class number in longer training}
Here we show results by training the model for 800 epochs on CUB, demonstrating the convergence after longer training. 

\begin{table}[h]
    \centering
    \footnotesize
    \begin{tabular}{cccccccccc}
    \toprule
    Epoch                & 200 & 400 & 600 & 700 & 720 & 740 & 760 & 780 & 800 \\
    \midrule
    GT $K^n=100$     &  112   &  122   & 114    &  108   &  109   &  107   &  108   &  106   & 107  \\
    \bottomrule
    \end{tabular}
    \caption{Results of estimated class number in longer training.}
    \label{tab:long_train}
\end{table}

}

\clearpage
\section{Error bars for generalized category discovery performance}
We repeatedly run our method and the previous state-of-the-art three times with different random seeds to show the mean and standard deviation values in~\cref{tab:generic_supp} and~\cref{tab:fine_grained_supp}, for both known and unknown class number cases. We can see that the variation is relatively small for all methods, and our method consistently outperforms the previous state-of-the-art across the board for both known and unknown class number cases.

\begin{table}[ht]
\small
\centering
\caption{\textbf{Results on generic image classification datasets.}}\label{tab:generic_supp}
\resizebox{\linewidth}{!}{ 
\begin{tabular}{clccccccccccc}
\toprule
&&     &   & \multicolumn{3}{c}{CIFAR10} & \multicolumn{3}{c}{CIFAR100} & \multicolumn{3}{c}{ImageNet-100}\\
\cmidrule(rl){5-7}
\cmidrule(rl){8-10}
\cmidrule(rl){11-13}
No. &Methods                       & \makecell{Known\\$K$} & PCA & All  & Old  & New  & All  & Old  & New  & All  & Old  & New \\\midrule
(1) & Vaze~\etal~\citet{vaze22generalized}      & \cmark & \xmark & 91.5$\pm$0.4 & 97.9$\pm$0.2 & 88.2$\pm$0.6 & 76.9$\pm$0.3 & 84.6$\pm$0.3 & 61.5$\pm$0.2 & 75.0$\pm$0.3 & 92.1$\pm$0.2 & 66.6$\pm$0.4 \\
\midrule
(2) & \textit{Ours} (GPC)                      & \cmark & \xmark & 91.9$\pm$0.2 & 98.2$\pm$0.3 & 88.6$\pm$0.1 & 77.6$\pm$0.4 & 84.9$\pm$0.4 & 62.7$\pm$0.4 & 76.7$\pm$0.4 & 94.3$\pm$0.2 & 68.8$\pm$0.3 \\ 
(3) & \textit{Ours} (GPC)                      & \cmark & \cmark & \textbf{91.9$\pm$0.4} & \textbf{98.2$\pm$0.3} & \textbf{89.1$\pm$0.2} & \textbf{77.8$\pm$0.3} & \textbf{85.3$\pm$0.2} & \textbf{63.5$\pm$0.2} & \textbf{77.3$\pm$0.4} & \textbf{94.6$\pm$0.4} & \textbf{71.1$\pm$0.3} \\
\midrule
(4) & Vaze~\etal~\citet{vaze22generalized}      & \xmark & \xmark & 88.6$\pm$0.5 & 96.2$\pm$0.4 & 84.9$\pm$0.6 & 73.2$\pm$0.4 & 83.5$\pm$0.4 & 57.9$\pm$0.4 & 72.7$\pm$0.4 & 91.8$\pm$0.5 & 63.8$\pm$0.6 \\
(5) & Vaze~\etal~\citet{vaze22generalized}      & \xmark & \cmark & 89.7$\pm$0.4 & 97.3$\pm$0.5 & 86.3$\pm$0.4 & 74.8$\pm$0.5 & 83.8$\pm$0.4 & 58.7$\pm$0.6 & 73.8$\pm$0.4 & 92.1$\pm$0.5 & 64.6$\pm$0.6 \\
\midrule
(6) & \textit{Ours} (GPC)                      & \xmark & \xmark & 88.2$\pm$0.4 & 97.0$\pm$0.5 & 85.9$\pm$0.3 & 75.1$\pm$0.5 & 84.4$\pm$0.4 & 59.9$\pm$0.6 & 74.9$\pm$0.5 & 93.2$\pm$0.4 & 65.5$\pm$0.3 \\
(7) & \textit{Ours} (GPC)                      & \xmark & \cmark & \textbf{90.6$\pm$0.3} & \textbf{98.2$\pm$0.4} & \textbf{87.1$\pm$0.4} & \textbf{75.7$\pm$0.5} & \textbf{84.7$\pm$0.6} & \textbf{60.9$\pm$0.4} & \textbf{75.7$\pm$0.3} & \textbf{93.4$\pm$0.4} & \textbf{66.8$\pm$0.5} \\
\bottomrule
\end{tabular}
}
\end{table}

\begin{table}[!ht]
\small
\centering
\caption{\textbf{Results on Semantic Shift Benchmark datasets.}}\label{tab:fine_grained_supp}
\resizebox{\linewidth}{!}{ 
\begin{tabular}{clccccccccccc}
\toprule
&&&& \multicolumn{3}{c}{CUB} & \multicolumn{3}{c}{Stanford Cars} & \multicolumn{3}{c}{FGVC-aircraft}\\
\cmidrule(rl){5-7}
\cmidrule(rl){8-10}
\cmidrule(rl){11-13}
No. & Methods                                  & \makecell{Known\\$K$} & PCA & All  & Old  & New  & All  & Old  & New  & All  & Old  & New \\
\midrule
(1) & Vaze~\etal~\citet{vaze22generalized}      & \cmark & \xmark & 51.1$\pm$0.2 & 56.4$\pm$0.1 & 48.4$\pm$0.3 & 39.1$\pm$0.3 & 57.6$\pm$0.4 & 29.9$\pm$0.3 & 45.1$\pm$0.2 & 41.2$\pm$0.3 & 46.8$\pm$0.2 \\
\midrule
(2) & \textit{Ours} (GPC)                      & \cmark & \xmark & 54.5$\pm$0.2 & 54.6$\pm$0.4 & 50.3$\pm$0.2 & 42.0$\pm$0.2 & 58.9$\pm$0.2 & 32.0$\pm$0.3 & 46.3$\pm$0.2 & 42.3$\pm$0.2 & 47.1$\pm$0.3 \\ 
(3) & \textit{Ours} (GPC)                      & \cmark & \cmark & \textbf{55.3$\pm$0.4} & \textbf{58.1$\pm$0.3} & \textbf{53.2$\pm$0.4} & \textbf{42.7$\pm$0.3} & \textbf{60.0$\pm$0.4} & \textbf{33.0$\pm$0.2} & \textbf{46.5$\pm$0.3} & \textbf{42.8$\pm$0.5} & \textbf{47.2$\pm$0.1} \\
\midrule
(4) & Vaze~\etal~\citet{vaze22generalized}      & \xmark & \xmark & 47.2$\pm$0.4 & 55.1$\pm$0.3 & 44.8$\pm$0.2 & 35.0$\pm$0.3 & 56.0$\pm$0.4 & 24.8$\pm$0.3 & 40.1$\pm$0.2 & 40.8$\pm$0.4 & 42.8$\pm$0.1 \\
(5) & Vaze~\etal~\citet{vaze22generalized}      & \xmark & \cmark & 49.2$\pm$0.3 & \textbf{56.2$\pm$0.2} & \textbf{46.3$\pm$0.4} & 36.3$\pm$0.3 & 56.6$\pm$0.4 & 25.9$\pm$0.5 & 41.2$\pm$0.3 & \textbf{40.9$\pm$0.4} & 44.6$\pm$0.2 \\
\midrule
(6) & \textit{Ours} (GPC)                      & \xmark & \xmark & 50.5$\pm$0.3 & 52.5$\pm$0.4 & 45.8$\pm$0.5 & 37.0$\pm$0.6 & 56.6$\pm$0.3 & 26.1$\pm$0.2 & 39.8$\pm$0.3 & 39.7$\pm$0.2 & 42.5$\pm$0.2 \\
(7) &\textit{Ours} (GPC)                       & \xmark & \cmark & \textbf{52.1$\pm$0.3} & 55.4$\pm$0.2 & 45.7$\pm$0.3 & \textbf{38.9$\pm$0.4} & \textbf{58.9$\pm$0.3} & \textbf{28.6$\pm$0.5} & \textbf{43.4$\pm$0.3} & 40.8$\pm$0.4 & \textbf{44.7$\pm$0.3} \\
\bottomrule
\end{tabular}
}
\end{table}

\iclrrevision{
\section{Error bars for category number estimation}
In this section, we show the estimated category numbers' standard deviations by repeatedly running our method with different random seeds.
The results are shown in~\cref{tab:number_error}. We can see that our method can estimate a more accurate category number to Vaze~\etal~\citet{vaze22generalized}.

\begin{table}[h]
    \centering
    \caption{Estimated category numbers}
    \label{tab:number_error}
    \begin{tabular}{lccccc}
    \toprule
    Estimated $K^n$          & CIFAR-10  & CIFAR-100 & ImageNet-100 & CUB-200 & Stanford-Cars \\
    \midrule
    \textit{Ours} (GPC)      &   5$\pm$1.4 &  22$\pm$2.6 &  54$\pm$3.1    &110$\pm$4.2   &   104$\pm$3.4    \\
    Vaze~\etal~\citet{vaze22generalized} &  4$\pm$1.2 & 23$\pm$1.5 &  59$\pm$4.3    &  131$\pm$5.6   &  132$\pm$2.5     \\             
    \midrule
    Ground Truth             & 5        & 20 & 50 & 100 & 98 \\
    \bottomrule
    \end{tabular}
\end{table}

}






\clearpage
\iclrrevision{

\section{Further comparison with ORCA}
ORCA~\citep{cao21orca} is originally pretrained only on the target dataset $\mathcal{D}$, \ie,  the data that our model is trained on.
We have shown the comparison using ImageNet pretrained features from DINO~\citep{caron2021emerging} for both ORCA~\citep{cao21orca} and our method in the main paper.
In~\cref{tab:orca_compare_generic} and~\cref{tab:orca_compare_ssb}, we provide additional comparison with ORCA, showing the effects of pretrained models using different data.


\begin{table}[h]
\small
\centering
\caption{\textbf{Comparison with ORCA~\citep{cao21orca} on generic classification datasets.}}\label{tab:orca_compare_generic}
\resizebox{0.8\linewidth}{!}{ 
\begin{tabular}{clcccccccccc}
\toprule
&     &   & \multicolumn{3}{c}{CIFAR10} & \multicolumn{3}{c}{CIFAR100} & \multicolumn{3}{c}{ImageNet-100}\\
\cmidrule(rl){4-6}
\cmidrule(rl){7-9}
\cmidrule(rl){10-12}
No. &Methods                       & Pretrain & All  & Old  & New  & All  & Old  & New  & All  & Old  & New \\
\midrule
(1) & ORCA~\citep{cao21orca}        & ImageNet & 91.4 & 88.0 & \textbf{91.2} & 68.9 & 76.1 & 46.6 & \textbf{79.8} & 93.6 & \textbf{74.9} \\
(2) & \textit{Ours} (GPC)           & ImageNet & \textbf{92.0} & \textbf{98.3} & 88.7 & \textbf{77.4} & \textbf{84.8} & \textbf{62.4} & 76.5 & \textbf{94.0} & 68.5 \\ 
\midrule
(3) & ORCA~\citep{cao21orca}        & Target   & 90.6 & 87.2 & 90.1 & 64.7 & 73.2 & 42.1 & 78.7 & \textbf{93.4} & 72.4 \\
(4) & \textit{Ours} (GPC)          & Target   & \textbf{91.1} & \textbf{87.8} & \textbf{90.5} & \textbf{65.0} & \textbf{74.3} & \textbf{42.6} & \textbf{79.6} & 93.3 & \textbf{73.1} \\ 
\bottomrule
\end{tabular}
}
\end{table}

\begin{table}[h]
\small
\centering
\caption{\textbf{Comparison with ORCA~\citep{cao21orca} on SSB~\citep{vaze22openset}.}}\label{tab:orca_compare_ssb}
\resizebox{0.8\linewidth}{!}{ 
\begin{tabular}{clcccccccccc}
\toprule
&     &   & \multicolumn{3}{c}{CUB} & \multicolumn{3}{c}{SCars} & \multicolumn{3}{c}{FGVC-Aircraft}\\
\cmidrule(rl){4-6}
\cmidrule(rl){7-9}
\cmidrule(rl){10-12}
No. &Methods                       & Pretrain & All  & Old  & New  & All  & Old  & New  & All  & Old  & New \\
\midrule
(1) & ORCA~\citep{cao21orca}        & ImageNet & 45.2 & \textbf{57.2} & 29.7 & 37.0 & \textbf{68.2} & 22.6 & \textbf{47.1} & \textbf{45.3} & 42.3 \\
(2) & \textit{Ours} (GPC)          & ImageNet & \textbf{54.2} & 54.9 & \textbf{50.3} & \textbf{41.2} & 58.8 & \textbf{31.6} & 46.1 & 42.4 & \textbf{47.2} \\ 
\midrule
(3) & ORCA~\citep{cao21orca}        & Target   & 42.9 & 52.0 & 28.4 & 40.3 & 57.0 & 31.4 & 44.4 & 40.7 & 44.1 \\
(4) & \textit{Ours} (GPC)          & Target   & \textbf{45.0} & \textbf{54.2} & \textbf{29.1} & \textbf{41.2} & \textbf{57.1} & \textbf{32.1} & \textbf{46.2} & \textbf{41.0} & \textbf{45.2} \\ 
\bottomrule
\end{tabular}
}
\end{table}

}

\clearpage
\section{Qualitative results}

\begin{figure*}[ht]
    \centering
    \includegraphics[width=0.8\textwidth]{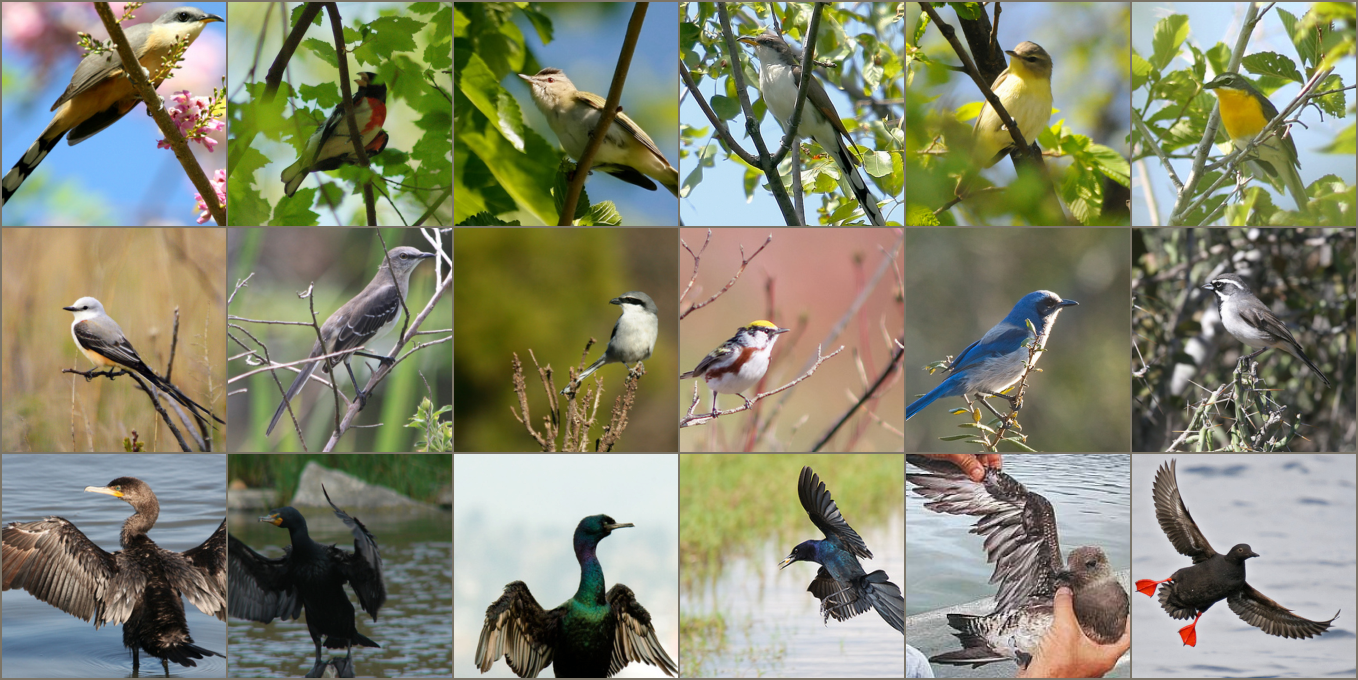}
    \caption{$k$-means grouping of features of DINO~\citep{caron2021emerging} on CUB-200 dataset. Notice that the grouping are roughly based on object pose or background, but we would want the clustering to be done to discriminate between different species. The kNN images to the randomly picked prototype (\ie, cluster center) are shown, from left (nearest) to right (furthest).}
    \label{fig:dino_prototype}
\end{figure*}

\begin{figure*}[ht]
    \centering
    \includegraphics[width=0.8\textwidth]{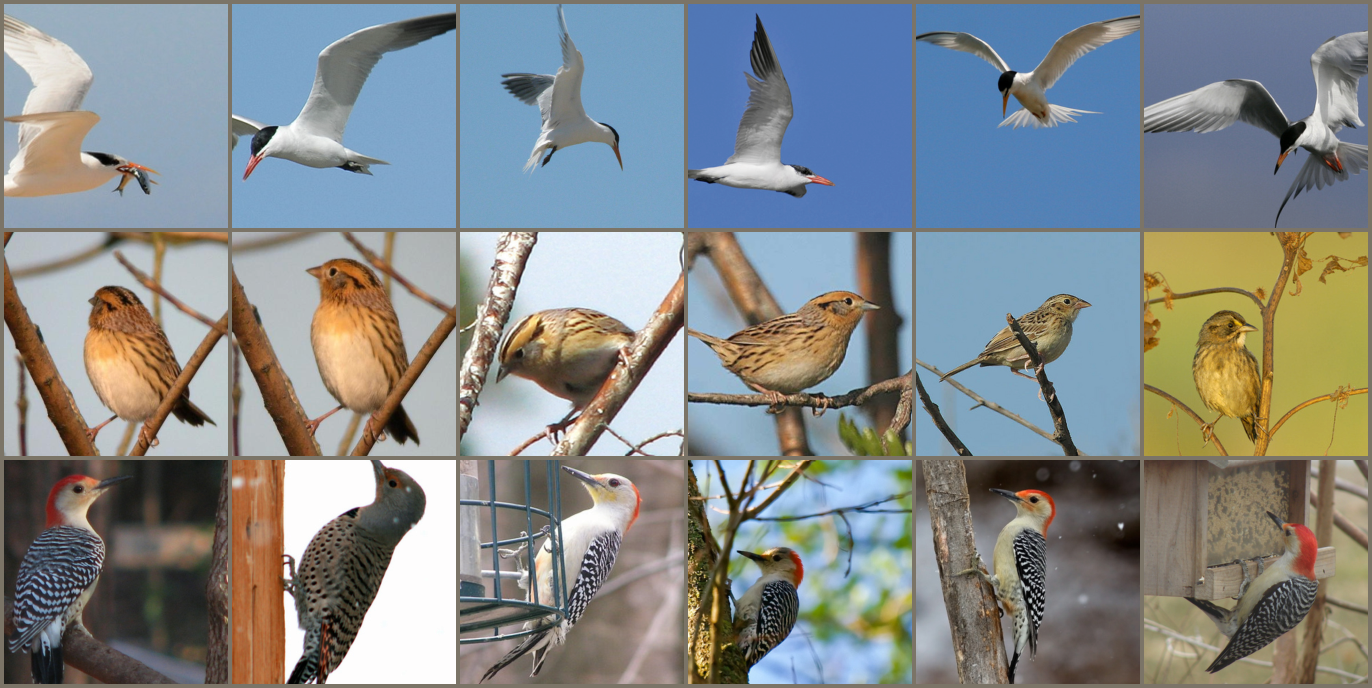}
    \caption{The prototype (\ie, the Gaussian mean vector in our method) and the retrieved nearest neighbor on GPC representations in the CUB-200 dataset. Images are grouped by different bird species. The kNN images to the randomly picked prototype (\ie, cluster center) are shown, from left (nearest) to right (furthest).}
    \label{fig:gpc_prototype}
\end{figure*}

In this section, we provide the visualization of the images grouped using the DINO features and our GPC trained features.
The results are presented in each row of~\cref{fig:dino_prototype} and~\cref{fig:gpc_prototype}.
The DINO features are effective to some extent when grouping images, but the results are still not satisfactory as the features are not tuned on the downstream tasks with a clear objective (see~\cref{fig:dino_prototype}).
On the other hand, after tuning the representation using our method, images from the same category can be grouped together (see~\cref{fig:gpc_prototype}).

\begin{figure}[!ht]
    \centering
    \includegraphics[width=\textwidth]{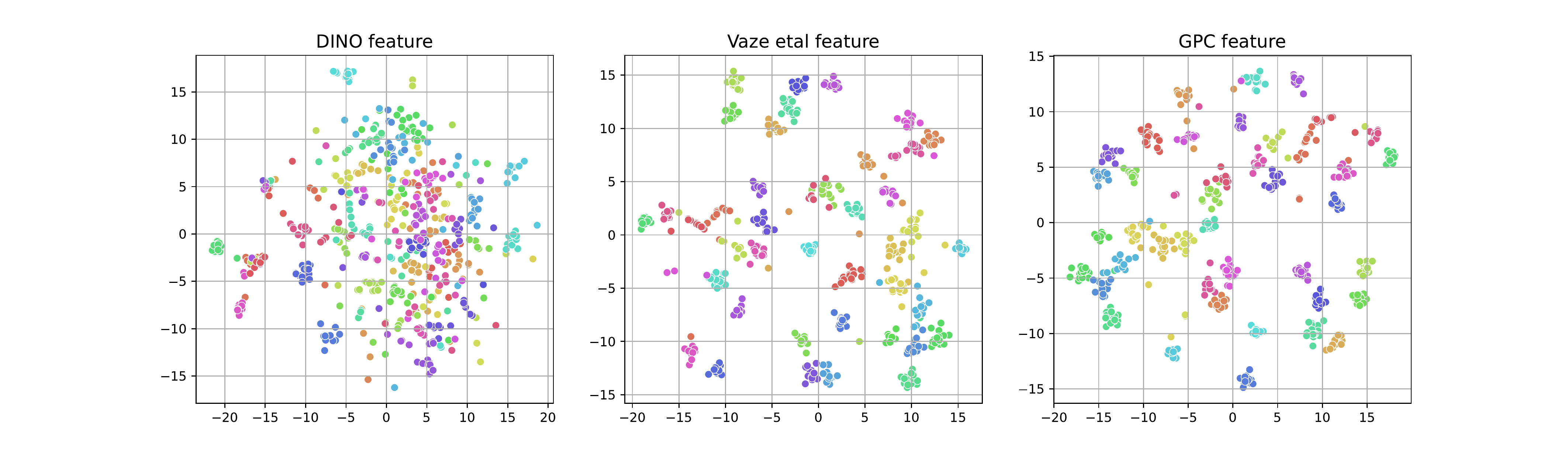}
    \caption{The t-SNE plot of the features on the CUB-200 dataset.}
    \label{fig:cub_tsne}
\end{figure}

\begin{figure}[ht]
    \centering
    \includegraphics[width=\textwidth]{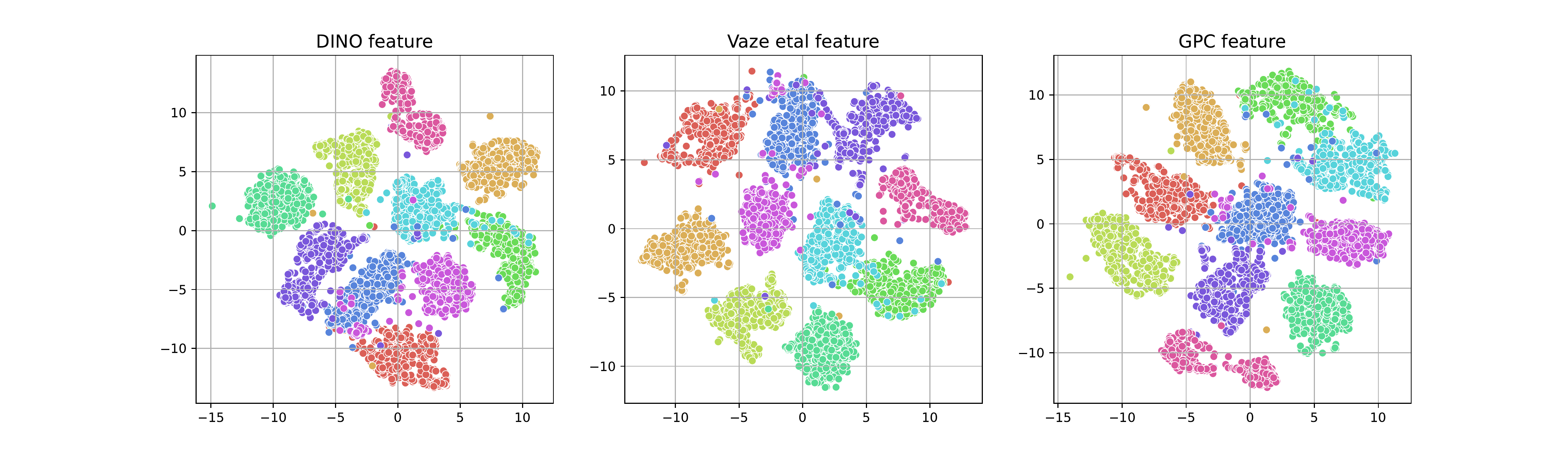}
    \caption{The t-SNE plot of the features on the ImageNet-100 dataset.}
    \label{fig:im100_tsne}
\end{figure}

\revision{We also present t-SNE projections of the learned features on both CUB-200 and ImageNet-100 datasets. From~\cref{fig:cub_tsne} we can see that on CUB-200, the DINO features can not separate different categories very well, while our method and \citet{vaze22generalized} can have a clear category boundary. From~\cref{fig:im100_tsne}, we found that although the t-SNE projections on ImageNet-100 appear to be similarly discriminative among DINO, \citet{vaze22generalized}, and our method, while further finetuning the representation from DINO can significantly improve the performance for the task of GCD.}

\clearpage
\section{Limitation and negative societal impact}
It should noted that although our method achieves the state-of-the-art results on the task of generalized category discovery, the classification performance is still far from those models trained with full human supervision.
Furthermore, when the class number  is unknown, there is still a noticeable performance gap \wrt\ the unknown category number case.
Besides, real-world data is much more complex and difficult than the curated data we used.
Therefore, careful validation and adaptation to specific application scenarios should be tested before deploying the model for any real-world use.

\section{License of used datasets}
All the datasets used in this paper are permitted for research use.
CIFAR-10 and CIFAR-100 datasets~\citep{cifar} are released under the MIT license, allowing use for research purposes.
The terms of access of the ImageNet dataset~\citep{deng2009imagenet} allow the use for non-commercial research and educational purposes. 
Similar to ImageNet, the Stanford Cars~\citep{stanfordcars} allows the use for research purposes.
The FGVC aircraft~\citep{aircraft} dataset was made available exclusively for non-commercial research purposes by the authors.
The CUB-200~\citep{cub200} dataset also allows research use.


%% file: cvpr2023_main.bbl
\begin{thebibliography}{10}\itemsep=-1pt

\bibitem{berthelot2019mixmatch}
David Berthelot, Nicholas Carlini, Ian Goodfellow, Nicolas Papernot, Avital
  Oliver, and Colin Raffel.
\newblock Mixmatch: A holistic approach to semi-supervised learning.
\newblock In {\em NeurIPS}, 2019.

\bibitem{cao21orca}
Kaidi Cao, Maria Brbi\'c, and Jure Leskovec.
\newblock Open-world semi-supervised learning.
\newblock In {\em ICLR}, 2022.

\bibitem{caron2021emerging}
Mathilde Caron, Hugo Touvron, Ishan Misra, Herv\'e J\'egou, Julien Mairal,
  Piotr Bojanowski, and Armand Joulin.
\newblock Emerging properties in self-supervised vision transformers.
\newblock In {\em ICCV}, 2021.

\bibitem{chen2020simple}
Ting Chen, Simon Kornblith, Mohammad Norouzi, and Geoffrey Hinton.
\newblock A simple framework for contrastive learning of visual
  representations.
\newblock In {\em ICML}, 2020.

\bibitem{chen2020improved}
Xinlei Chen, Haoqi Fan, Ross Girshick, and Kaiming He.
\newblock Improved baselines with momentum contrastive learning.
\newblock {\em arXiv preprint arXiv:2003.04297}, 2020.

\bibitem{comaniciu2002mean_shift}
Dorin Comaniciu and Peter Meer.
\newblock Mean shift: A robust approach toward feature space analysis.
\newblock {\em IEEE TPAMI}, 2002.

\bibitem{cui2022discriminability}
Quan Cui, Bingchen Zhao, Zhao-Min Chen, Borui Zhao, Renjie Song, Jiajun Liang,
  Boyan Zhou, and Osamu Yoshie.
\newblock Discriminability-transferability trade-off: An information-theoretic
  perspective.
\newblock In {\em ECCV}, 2022.

\bibitem{deng2009imagenet}
Jia Deng, Wei Dong, Richard Socher, Li-Jia Li, Kai Li, and Li Fei-Fei.
\newblock Imagenet: A large-scale hierarchical image database.
\newblock In {\em CVPR}, 2009.

\bibitem{dosovitskiy2020vit}
Alexey Dosovitskiy, Lucas Beyer, Alexander Kolesnikov, Dirk Weissenborn,
  Xiaohua Zhai, Thomas Unterthiner, Mostafa Dehghani, Matthias Minderer, Georg
  Heigold, Sylvain Gelly, Jakob Uszkoreit, and Neil Houlsby.
\newblock An image is worth 16x16 words: Transformers for image recognition at
  scale.
\newblock In {\em ICLR}, 2021.

\bibitem{ester1996density}
Martin Ester, Hans-Peter Kriegel, J{\"o}rg Sander, Xiaowei Xu, et~al.
\newblock A density-based algorithm for discovering clusters in large spatial
  databases with noise.
\newblock In {\em KDD}, 1996.

\bibitem{fei2022xcon}
Yixin Fei, Zhongkai Zhao, Siwei Yang, and Bingchen Zhao.
\newblock Xcon: Learning with experts for fine-grained category discovery.
\newblock In {\em BMVC}, 2022.

\bibitem{fini2021unified}
Enrico Fini, Enver Sangineto, St{\'e}phane Lathuili{\`e}re, Zhun Zhong, Moin
  Nabi, and Elisa Ricci.
\newblock A unified objective for novel class discovery.
\newblock In {\em ICCV}, 2021.

\bibitem{ghasedi2017deep}
Kamran Ghasedi~Dizaji, Amirhossein Herandi, Cheng Deng, Weidong Cai, and Heng
  Huang.
\newblock Deep clustering via joint convolutional autoencoder embedding and
  relative entropy minimization.
\newblock In {\em ICCV}, 2017.

\bibitem{rankstat}
Kai Han, Sylvestre-Alvise Rebuffi, Sebastien Ehrhardt, Andrea Vedaldi, and
  Andrew Zisserman.
\newblock Automatically discovering and learning new visual categories with
  ranking statistics.
\newblock In {\em ICLR}, 2020.

\bibitem{han21autonovel}
Kai Han, Sylvestre-Alvise Rebuffi, Sebastien Ehrhardt, Andrea Vedaldi, and
  Andrew Zisserman.
\newblock Autonovel: Automatically discovering and learning novel visual
  categories.
\newblock {\em IEEE TPAMI}, 2021.

\bibitem{Han2019learning}
Kai Han, Andrea Vedaldi, and Andrew Zisserman.
\newblock Learning to discover novel visual categories via deep transfer
  clustering.
\newblock In {\em ICCV}, 2019.

\bibitem{hastings1970monte}
Wilfred~Keith Hastings.
\newblock Monte carlo sampling methods using markov chains and their
  applications.
\newblock {\em Biometrika}, 1970.

\bibitem{he2019moco}
Kaiming He, Haoqi Fan, Yuxin Wu, Saining Xie, and Ross Girshick.
\newblock Momentum contrast for unsupervised visual representation learning.
\newblock In {\em CVPR}, 2020.

\bibitem{resnet}
Kaiming He, Xiangyu Zhang, Shaoqing Ren, and Jian Sun.
\newblock Deep residual learning for image recognition.
\newblock In {\em CVPR}, 2016.

\bibitem{hsu2017learning}
Yen-Chang Hsu, Zhaoyang Lv, and Zsolt Kira.
\newblock Learning to cluster in order to transfer across domains and tasks.
\newblock In {\em ICLR}, 2018.

\bibitem{hsu2019multi}
Yen-Chang Hsu, Zhaoyang Lv, Joel Schlosser, Phillip Odom, and Zsolt Kira.
\newblock Multi-class classification without multi-class labels.
\newblock In {\em ICLR}, 2019.

\bibitem{hua2021feature}
Tianyu Hua, Wenxiao Wang, Zihui Xue, Yue Wang, Sucheng Ren, and Hang Zhao.
\newblock On feature decorrelation in self-supervised learning.
\newblock In {\em ICCV}, 2021.

\bibitem{huang2021trash}
Junkai Huang, Chaowei Fang, Weikai Chen, Zhenhua Chai, Xiaolin Wei, Pengxu Wei,
  Liang Lin, and Guanbin Li.
\newblock Trash to treasure: harvesting ood data with cross-modal matching for
  open-set semi-supervised learning.
\newblock In {\em ICCV}, 2021.

\bibitem{jia2021joint}
Xuihui Jia, Kai Han, Yukun Zhu, and Bradley Green.
\newblock Joint representation learning and novel category discovery on
  single-and multi-modal data.
\newblock In {\em ICCV}, 2021.

\bibitem{Jing2021UnderstandingDC}
Li Jing, Pascal Vincent, Yann LeCun, and Yuandong Tian.
\newblock Understanding dimensional collapse in contrastive self-supervised
  learning.
\newblock In {\em ICLR}, 2021.

\bibitem{stanfordcars}
Jonathan Krause, Michael Stark, Jia Deng, and Li Fei-Fei.
\newblock 3d object representations for fine-grained categorization.
\newblock In {\em 4th International IEEE Workshop on 3D Representation and
  Recognition (3dRR-13)}, 2013.

\bibitem{cifar}
Alex Krizhevsky and Geoffrey Hinton.
\newblock Learning multiple layers of features from tiny images.
\newblock {\em Technical Report}, 2009.

\bibitem{laine2016temporal}
Samuli Laine and Timo Aila.
\newblock Temporal ensembling for semi-supervised learning.
\newblock In {\em ICLR}, 2017.

\bibitem{li2020prototypical}
Junnan Li, Pan Zhou, Caiming Xiong, and Steven~CH Hoi.
\newblock Prototypical contrastive learning of unsupervised representations.
\newblock In {\em ICLR}, 2020.

\bibitem{liu2023large}
Mingxuan Liu, Subhankar Roy, Zhun Zhong, Nicu Sebe, and Elisa Ricci.
\newblock Large-scale pre-trained models are surprisingly strong in incremental
  novel class discovery.
\newblock {\em arXiv preprint arXiv:2303.15975}, 2023.

\bibitem{macqueen1967some_kmeans}
James MacQueen.
\newblock Some methods for classification and analysis of multivariate
  observations.
\newblock In {\em Proceedings of the Fifth Berkeley Symposium on Mathematical
  Statistics and Probability}, 1967.

\bibitem{aircraft}
Subhransu Maji, Esa Rahtu, Juho Kannala, Matthew Blaschko, and Andrea Vedaldi.
\newblock Fine-grained visual classification of aircraft.
\newblock {\em arXiv preprint arXiv:1306.5151}, 2013.

\bibitem{murtagh2014ward}
Fionn Murtagh and Pierre Legendre.
\newblock Ward’s hierarchical agglomerative clustering method: which
  algorithms implement ward’s criterion?
\newblock {\em Journal of classification}, 2014.

\bibitem{oliver2018realistic}
Avital Oliver, Augustus Odena, Colin Raffel, Ekin~D Cubuk, and Ian~J
  Goodfellow.
\newblock Realistic evaluation of deep semi-supervised learning algorithms.
\newblock In {\em NeurIPS}, 2018.

\bibitem{Pu_2023_CVPR}
Nan Pu, Zhun Zhong, and Nicu Sebe.
\newblock Dynamic conceptional contrastive learning for generalized category
  discovery.
\newblock In {\em CVPR}, 2023.

\bibitem{rebuffi2020semi}
Sylvestre-Alvise Rebuffi, Sebastien Ehrhardt, Kai Han, Andrea Vedaldi, and
  Andrew Zisserman.
\newblock Semi-supervised learning with scarce annotations.
\newblock In {\em CVPR Deep-Vision workshop}, 2020.

\bibitem{rebuffi21lsdc}
Sylvestre-Alvise Rebuffi, Sebastien Ehrhardt, Kai Han, Andrea Vedaldi, and
  Andrew Zisserman.
\newblock Lsd-c: Linearly separable deep clusters.
\newblock In {\em ICCV Workshop on Visual Inductive Priors for Data-Efficient
  Deep Learning}, 2021.

\bibitem{ronen2022deepdpm}
Meitar Ronen, Shahaf~E. Finder, and Oren Freifeld.
\newblock Deepdpm: Deep clustering with an unknown number of clusters.
\newblock In {\em CVPR}, 2022.

\bibitem{incd2022}
Subhankar Roy, Mingxuan Liu, Zhun Zhong, Nicu Sebe, and Elisa Ricci.
\newblock Class-incremental novel class discovery.
\newblock In {\em ECCV}, 2022.

\bibitem{saito2021openmatch}
Kuniaki Saito, Donghyun Kim, and Kate Saenko.
\newblock Openmatch: Open-set consistency regularization for semi-supervised
  learning with outliers.
\newblock In {\em NeurIPS}, 2021.

\bibitem{sohn2020fixmatch}
Kihyuk Sohn, David Berthelot, Chun-Liang Li, Zizhao Zhang, Nicholas Carlini,
  Ekin~D Cubuk, Alex Kurakin, Han Zhang, and Colin Raffel.
\newblock Fixmatch: Simplifying semi-supervised learning with consistency and
  confidence.
\newblock In {\em NeurIPS}, 2020.

\bibitem{sun2023opencon}
Yiyou Sun and Yixuan Li.
\newblock Opencon: Open-world contrastive learning.
\newblock {\em TMLR}, 2023.

\bibitem{tarvainen2017mean}
Antti Tarvainen and Harri Valpola.
\newblock Mean teachers are better role models: Weight-averaged consistency
  targets improve semi-supervised deep learning results.
\newblock In {\em NeurIPS}, 2017.

\bibitem{cmc}
Yonglong Tian, Dilip Krishnan, and Phillip Isola.
\newblock Contrastive multiview coding.
\newblock In {\em ECCV}, 2020.

\bibitem{vaze22generalized}
Sagar Vaze, Kai Han, Andrea Vedaldi, and Andrew Zisserman.
\newblock Generalized category discovery.
\newblock In {\em CVPR}, 2022.

\bibitem{vaze22openset}
Sagar Vaze, Kai Han, Andrea Vedaldi, and Andrew Zisserman.
\newblock Open-set recognition: A good closed-set classifier is all you need?
\newblock In {\em ICLR}, 2022.

\bibitem{cub200}
Catherine Wah, Steve Branson, Peter Welinder, Pietro Perona, and Serge
  Belongie.
\newblock {Caltech-UCSD Birds 200}.
\newblock {\em Computation \& Neural Systems Technical Report}, 2010.

\bibitem{wen2022simple}
Xin Wen, Bingchen Zhao, and Xiaojuan Qi.
\newblock Parametric classification for generalized category discovery: A
  baseline study.
\newblock In {\em ICCV}, 2023.

\bibitem{wu2018unsupervised}
Zhirong Wu, Yuanjun Xiong, Stella~X Yu, and Dahua Lin.
\newblock Unsupervised feature learning via non-parametric instance
  discrimination.
\newblock In {\em CVPR}, 2018.

\bibitem{xie2016unsupervised}
Junyuan Xie, Ross Girshick, and Ali Farhadi.
\newblock Unsupervised deep embedding for clustering analysis.
\newblock In {\em ICML}, 2016.

\bibitem{yu2020multi}
Qing Yu, Daiki Ikami, Go Irie, and Kiyoharu Aizawa.
\newblock Multi-task curriculum framework for open-set semi-supervised
  learning.
\newblock In {\em ECCV}, 2020.

\bibitem{zhai2019s4l}
Xiaohua Zhai, Avital Oliver, Alexander Kolesnikov, and Lucas Beyer.
\newblock S4l: Self-supervised semi-supervised learning.
\newblock In {\em ICCV}, 2019.

\bibitem{zhang2022promptcal}
Sheng Zhang, Salman Khan, Zhiqiang Shen, Muzammal Naseer, Guangyi Chen, and
  Fahad Khan.
\newblock Promptcal: Contrastive affinity learning via auxiliary prompts for
  generalized novel category discovery.
\newblock In {\em CVPR}, 2023.

\bibitem{zhao21novel}
Bingchen Zhao and Kai Han.
\newblock Novel visual category discovery with dual ranking statistics and
  mutual knowledge distillation.
\newblock In {\em NeurIPS}, 2021.

\bibitem{zhao2023incremental}
Bingchen Zhao and Oisin Mac~Aodha.
\newblock Incremental generalized category discovery.
\newblock In {\em ICCV}, 2023.

\bibitem{zhong2021neighborhood}
Zhun Zhong, Enrico Fini, Subhankar Roy, Zhiming Luo, Elisa Ricci, and Nicu
  Sebe.
\newblock Neighborhood contrastive learning for novel class discovery.
\newblock In {\em CVPR}, 2021.

\bibitem{zhong2021openmix}
Zhun Zhong, Linchao Zhu, Zhiming Luo, Shaozi Li, Yi Yang, and Nicu Sebe.
\newblock Openmix: Reviving known knowledge for discovering novel visual
  categories in an open world.
\newblock In {\em CVPR}, 2021.

\bibitem{zhu2021improving}
Rui Zhu, Bingchen Zhao, Jingen Liu, Zhenglong Sun, and Chang~Wen Chen.
\newblock Improving contrastive learning by visualizing feature transformation.
\newblock In {\em ICCV}, 2021.

\end{thebibliography}
